\documentclass[11pt,a4paper,preprint, authoryear]{elsarticle}

\usepackage{makecell}
\usepackage{rotating}
\usepackage{adjustbox}
\usepackage{lscape}

\newlength{\cslhangindent}
\newenvironment{CSLReferences}%
\usepackage{lmodern}
\usepackage{setspace}
\DeclareMathSizes{12}{14}{10}{10}

\usepackage{float}
\let\origfigure\figure
\let\endorigfigure\endfigure
\renewenvironment{figure}[1][2] {
    \expandafter\origfigure\expandafter[H]
} {
    \endorigfigure
}

\let\origtable\table
\let\endorigtable\endtable
\renewenvironment{table}[1][2] {
    \expandafter\origtable\expandafter[H]
} {
    \endorigtable
}

\usepackage{ifxetex,ifluatex}
\usepackage{fixltx2e} % provides \textsubscript
\ifnum 0\ifxetex 1\fi\ifluatex 1\fi=0 % if pdftex
  \usepackage[T1]{fontenc}
  \usepackage[utf8]{inputenc}
\else % if luatex or xelatex
  \ifxetex
    \usepackage{mathspec}
    \usepackage{xltxtra,xunicode}
  \else
    \usepackage{fontspec}
  \fi
  \defaultfontfeatures{Mapping=tex-text,Scale=MatchLowercase}
  
\fi

\usepackage{amssymb, amsmath, amsthm, amsfonts}
\theoremstyle{plain}

\newtheorem{proposition}{Proposition}

\usepackage[round]{natbib}

\usepackage{longtable}
\usepackage[margin=2.3cm,bottom=2cm,top=2.5cm, includefoot]{geometry}
\usepackage{fancyhdr}
\usepackage[bottom, hang, flushmargin]{footmisc}
\usepackage{graphicx}
\numberwithin{equation}{section}
\numberwithin{figure}{section}
\numberwithin{table}{section}
\usepackage{textcomp}

\usepackage{array}
\newcolumntype{x}[1]{>{\centering\arraybackslash\hspace{0pt}}p{#1}}

\makeatletter
\def\ps@pprintTitle{%
  \let\@oddhead\@empty
  \let\@evenhead\@empty
  \let\@oddfoot\@empty
  \let\@evenfoot\@oddfoot
}
\makeatother

\usepackage{hyperref}
\hypersetup{breaklinks=true,
            bookmarks=true,
            colorlinks=true,
            citecolor=blue,
            urlcolor=blue,
            linkcolor=blue,
            pdfborder={0 0 0}}

\usepackage{siunitx}
\usepackage{multirow}
\usepackage{hhline}
\usepackage{calc}
\usepackage{tabularx}
\usepackage{booktabs}
\usepackage{caption}

\def\useignorespacesandallpars#1\ignorespaces\fi{%
#1\fi\ignorespacesandallpars}

\makeatletter
\def\ignorespacesandallpars{%
  \@ifnextchar\par
    {\expandafter\ignorespacesandallpars\@gobble}%
    {}%
}
\makeatother

\let\rmarkdownfootnote\footnote%
\def\footnote{\protect\rmarkdownfootnote}
\IfFileExists{upquote.sty}{\usepackage{upquote}}{}

\usepackage{booktabs}
\usepackage{longtable}
\usepackage{array}
\usepackage{multirow}
\usepackage{wrapfig}
\usepackage{float}
\usepackage{colortbl}
\usepackage{pdflscape}
\usepackage{tabu}
\usepackage{threeparttable}
\usepackage{threeparttablex}
\usepackage[normalem]{ulem}
\usepackage{makecell}
\usepackage{xcolor}

\begin{document}

\begin{frontmatter}  %

\title{Cluster Regularization via a Hierarchical Feature Regression}

% Set to FALSE if wanting to remove title (for submission)

\author[Add1]{Johann Pfitzinger\footnote{This paper represents a chapter
  of my PhD thesis submitted at Goethe University Frankfurt. I thank my
  supervisor Uwe Hassler for his advice.}}
\ead{johann.pfitzinger@gmail.com}

\address[Add1]{Goethe University, Frankfurt am Main}

\begin{abstract}
\small{
This paper proposes a novel graph-based regularized regression estimator
--- the hierarchical feature regression (HFR) ---, which mobilizes
insights from the domains of machine learning and graph theory to
estimate robust parameters for a linear regression. The estimator
constructs a supervised feature graph that decomposes parameters along
its edges, adjusting first for common variation and successively
incorporating idiosyncratic patterns into the fitting process. The graph
structure has the effect of shrinking parameters towards group targets,
where the extent of shrinkage is governed by a hyperparamter, and group
compositions as well as shrinkage targets are determined endogenously.
The method offers rich resources for the visual exploration of the
latent effect structure in the data, and demonstrates good predictive
accuracy and versatility when compared to a panel of commonly used
regularization techniques across a range of empirical and simulated
regression tasks.
}
\end{abstract}

\vspace{1cm}

\begin{keyword}
\footnotesize{
Regularized regression \sep group shrinkage \sep machine learning
\sep supervised hierarchical clustering \\ \vspace{0.3cm}
}
\end{keyword}
\vspace{0.5cm}
\end{frontmatter}

\newpage

%________________________
% Header and Footers
%%%%%%%%%%%%%%%%%%%%%%%%%%%%%%%%%
\pagestyle{fancy}
\chead{}
\rhead{}
\lfoot{}
\rfoot{\footnotesize Page \thepage}
\lhead{}
%\rfoot{\footnotesize Page \thepage } % "e.g. Page 2"
\cfoot{}

%\setlength\headheight{30pt}
%%%%%%%%%%%%%%%%%%%%%%%%%%%%%%%%%
%________________________

\headsep 35pt % So that header does not go over title

\hypertarget{introduction}{%
\section{\texorpdfstring{Introduction
\label{Introduction}}{Introduction }}\label{introduction}}

In this paper, I propose a new solution to the old problem of obtaining
robust parameter estimates in a high-dimensional regression with
nonorthogonal predictors. I decompose the estimates of an ordinary least
squares regression along a supervised hierarchical graph, then optimally
shrink the edges of the graph to achieve a group-wise regularization of
the parameter space. The resulting estimator has several useful
properties: (i) It solves the problem of group shrinkage in an elegant
and efficient manner, where the composition of parameter groups as well
as group shrinkage targets are determined endogenously; (ii) The
estimator offers intuitive tools for the visual inspection of the model
effects structure; (iii) It exhibits significant versatility, performing
well (in terms of prediction accuracy) both in sparse, as well as dense
regression settings; Finally, (iv) the estimator encodes the prior
expectation of a world governed by hierarchical processes, making it
uniquely suitable for several empirical applications, particularly in
the domains of economics and finance.

A substantial literature exists on regularized regression techniques,
the main thrust of which comprises variants of penalized or latent
variable regressions, and which finds its most general expression in the
extensive field of Bayesian regression analysis. With increasing
availability of data, regularized regressions have steadily grown in
importance in many fields, and underpin developments in domains as
seemingly disparate as bioinformatics, finance or deep learning.
Economic applications in particular are often characterized by
high-dimensional, multicollinear data sets, and regularized machine
learning algorithms are well established as computationally efficient
means of obtaining accurate parameter estimates when the number of
predictors relative to observations is high. The hierarchical feature
regression (HFR) contributes to this body of knowledge, combining
elements of graph theory and machine learning to inform a novel group
shrinkage estimator.

The HFR constructs a parsimonious information graph, using a supervised
hierarchical clustering algorithm that groups predictors based on the
similarity of their explanatory content with respect to a dependent
variable. The information graph is translated into a parameter
hierarchy, consisting of several chains of coefficients (edges in the
graph) that capture increasingly nuanced signal. The coefficient chains
adjust first for shared variation, with each lower element introducing a
further degree of idiosyncrasy. By shrinking the chain of coefficients,
the HFR achieves group shrinkage --- removing idiosyncratic information
from the fitting process and giving a higher weight to shared effect
patterns.

An economic case study highlights how the structure introduced by the
hierarchical graph can be exploited to garner insights into latent
effect dynamics in the fitted model, with rich resources for visual
exploration. Furthermore, the HFR exhibits robust predictive accuracy,
comparing favorably against a panel of benchmark regularized regression
techniques. The results also indicate a high degree of versatilty in the
simulated setting, with good performance across different types of
regression settings (e.g.~sparse, latent factors, grouped). This
flexibility is a key advantage: where related methods tend to be best
suited to specialized types of tasks, the HFR can produce accurate
parameter estimates across a spectrum of data generating processes.

The remainder of this paper is structured as follows: Section
\ref{literature} introduces important literature relating to the field
of regularized regression. The HFR is developed in Section \ref{method},
while Sections \ref{empirical} and \ref{simulations} explore its
performance both in empirical and simulated settings. Finally, Section
\ref{conclusion} concludes the paper.

\hypertarget{literature-review}{%
\section{\texorpdfstring{Literature review
\label{literature}}{Literature review }}\label{literature-review}}

Nobel prize laureate Herbert Simon posits that complex systems tend to
evolve in a hierarchic manner and, as a result, encompass hierarchical
structures
(\protect\hyperlink{ref-simonArchitectureComplexity1962}{Simon, 1962}).
This proposition is supported by an understanding of highly integrated
markets and economies driven in part by deeper global undercurrents ---
e.g.~global business cycles
(\protect\hyperlink{ref-dieboldMeasuringDynamicsGlobal2015}{Diebold \&
Yilmaz, 2015};
\protect\hyperlink{ref-koseInternationalBusinessCycles2003}{Kose
\emph{et al.}, 2003}) or global financial cycles
(\protect\hyperlink{ref-reyDilemmaNotTrilemma2015}{Rey, 2015}) ---, and
is reflected in the popularity of latent variable methods (e.g.~dynamic
factor models for macroeconometric analysis) and, increasingly, deep
learning methods for nonlinear prediction tasks.\footnote{Deep neural
  networks, for instance, have been described as nonlinear hierarchical
  feature methods
  (\protect\hyperlink{ref-mishraDeepMachineLearning2017}{Mishra \&
  Gupta, 2017}).}

The HFR utilizes empirical data hierarchies with the objective of
achieving an optimal group mean shrinkage that captures the hierarchical
nature of the data generating processes and, in turn, attains more
robust out-of-sample performance. It is therefore located squarely
within the regularization literature. A plethora of approaches to
parameter regularization have been developed in this domain. Penalized
regressions --- termed ``Lasso and friends'' in
\protect\hyperlink{ref-varianBigDataNew2014}{Varian}
(\protect\hyperlink{ref-varianBigDataNew2014}{2014}) --- receive some
attention in this paper as natural benchmarks for the HFR. The
approaches introduce a constraint on the parameter norm, by adding a
penalty function \(P_{\lambda}(\boldsymbol{\beta})\) to the least
squares loss of a regression of \(y\) on \(\mathbf{x}\):
\begin{equation}
\boldsymbol{\hat{\beta}} = \arg \min_{\boldsymbol{\beta}} \left[N^{-1} (y - \boldsymbol{x \beta})'(y - \boldsymbol{x \beta}) + P_{\lambda}(\boldsymbol{\beta})\right].
\end{equation} Here \(\boldsymbol{\hat{\beta}}\) is a vector of
parameter estimates and \(N\) is the sample size. The penalty function
depends on a hyperparameter \(\lambda\) governing the weight given to
the penalty, and typically takes the form
\(P_{\lambda}(\boldsymbol{\beta}) = \lambda \sum_i |\beta_i|^q\), where
\(q = 1\) is a Lasso and \(q = 2\) is a ridge regression. Important
contributions to this literature include
\protect\hyperlink{ref-jamesEstimationQuadraticLoss1961}{James \& Stein}
(\protect\hyperlink{ref-jamesEstimationQuadraticLoss1961}{1961}),
\protect\hyperlink{ref-hoerlApplicationRidgeAnalysis1962}{Hoerl}
(\protect\hyperlink{ref-hoerlApplicationRidgeAnalysis1962}{1962}),
\protect\hyperlink{ref-hoerlRidgeRegressionBiased1970}{Hoerl \& Kennard}
(\protect\hyperlink{ref-hoerlRidgeRegressionBiased1970}{1970}),
\protect\hyperlink{ref-tibshiraniRegressionShrinkageSelection1996}{Tibshirani}
(\protect\hyperlink{ref-tibshiraniRegressionShrinkageSelection1996}{1996})
and \protect\hyperlink{ref-efronLeastAngleRegression2004}{Efron \emph{et
al.}} (\protect\hyperlink{ref-efronLeastAngleRegression2004}{2004}), as
well as multiple variants, including
\protect\hyperlink{ref-zouRegularizationVariableSelection2005}{Zou \&
Hastie}
(\protect\hyperlink{ref-zouRegularizationVariableSelection2005}{2005}),
\protect\hyperlink{ref-zouAdaptiveLassoIts2006}{Zou}
(\protect\hyperlink{ref-zouAdaptiveLassoIts2006}{2006}) and
\protect\hyperlink{ref-zouAdaptiveElasticNetDiverging2009}{Zou \& Zhang}
(\protect\hyperlink{ref-zouAdaptiveElasticNetDiverging2009}{2009}). An
introductory overview is found in
\protect\hyperlink{ref-friedmanElementsStatisticalLearning2001}{Friedman
\emph{et al.}}
(\protect\hyperlink{ref-friedmanElementsStatisticalLearning2001}{2001}).

Penalized regressions --- particularly those based on the
\(\ell_1\)-norm (\(q = 1\)) --- have been extended to permit group
shrinkage
(\protect\hyperlink{ref-bondellSimultaneousRegressionShrinkage2008}{Bondell
\& Reich, 2008};
\protect\hyperlink{ref-tibshiraniSparsitySmoothnessFused2005}{Tibshirani
\emph{et al.}, 2005};
\protect\hyperlink{ref-turlachSimultaneousVariableSelection2005}{Turlach
\emph{et al.}, 2005};
\protect\hyperlink{ref-yuanModelSelectionEstimation2006}{Yuan \& Lin,
2006}; \protect\hyperlink{ref-zengNovelSparsityClustering2013}{Zeng \&
Figueiredo, 2013}). A good review of available approaches is given in
\protect\hyperlink{ref-bachStructuredSparsityConvex2012}{Bach \emph{et
al.}} (\protect\hyperlink{ref-bachStructuredSparsityConvex2012}{2012}).
Group shrinkage typically aims to shrink disjoint or overlapping groups
of variables towards zero, often requiring prior knowledge of groups.
The HFR differs from these methods in that sparsity is not an objective
and group compositions are estimated endogenously without the need for
external structures.

Conceptually, group shrinkage can be achieved in a penalized regression
framework, for instance, by generalizing the ridge regression to the
following form (\protect\hyperlink{ref-hansenEconometrics2019}{Hansen,
2019}; \protect\hyperlink{ref-vanwieringenLectureNotesRidge2020}{van
Wieringen, 2020}): \begin{equation}
\boldsymbol{\hat{\beta}} = \arg \min_{\beta} \left[ (y - \boldsymbol{x\beta})'(y - \boldsymbol{x\beta}) + (\boldsymbol{\beta}_0 - \boldsymbol{\beta})'\boldsymbol{\Delta}(\boldsymbol{\beta}_0 - \boldsymbol{\beta}) \right],
\end{equation} where \(\boldsymbol{\Delta}\) governs the speed and
direction of shrinkage for each parameter individually, and
\(\boldsymbol{\beta}_0\) contains a shrinkage target for each parameter.
The target values can be set in such a way as to induce group-wise
shrinkage, by selecting the same shrinkage target for groups of
variables, and specifying penalties in \(\boldsymbol{\Delta}\) on a
group-specific basis. This requires \emph{a priori} definitions of group
compositions and target values, reducing its practicality.

A second broad class of regularization techniques are latent variable
regressions. Examples include the principal components regression (PCR)
described in
\protect\hyperlink{ref-friedmanElementsStatisticalLearning2001}{Friedman
\emph{et al.}}
(\protect\hyperlink{ref-friedmanElementsStatisticalLearning2001}{2001}),
the partial least squares regression (PLSR) developed by Wold in the
1960s and 70s (see
\protect\hyperlink{ref-woldPersonalMemoriesEarly2001}{Wold}
(\protect\hyperlink{ref-woldPersonalMemoriesEarly2001}{2001}) and
\protect\hyperlink{ref-martensReliableRelevantModelling2001}{Martens}
(\protect\hyperlink{ref-martensReliableRelevantModelling2001}{2001})),
or --- in the econometric setting --- the dynamic factor model surveyed
in \protect\hyperlink{ref-stockDynamicFactorModels2016}{Stock \& Watson}
(\protect\hyperlink{ref-stockDynamicFactorModels2016}{2016a}) and
\protect\hyperlink{ref-stockFactorModelsStructural2016}{Stock \& Watson}
(\protect\hyperlink{ref-stockFactorModelsStructural2016}{2016b}). These
methods reduce the dimensionality of the predictor set by removing low
variance components in the case of principal components based methods,
or components with a low response correlation in the case of PLSR
(\protect\hyperlink{ref-jolliffePrincipalComponentAnalysis2002}{Jolliffe,
2002}). Unlike penalized regressions, latent variable regressions are
mostly unsupervised in their construction of latent factors. Some
exceptions exist, for instance the aforementioned PLSR, or
\protect\hyperlink{ref-bairPredictionSupervisedPrincipal2006}{Bair
\emph{et al.}}
(\protect\hyperlink{ref-bairPredictionSupervisedPrincipal2006}{2006}),
who introduce a (semi-)supervised PCR, by using a supervised process of
pre-filtering the predictor set before performing principal components
analysis.

The HFR constructs factors using a hierarchical transformation of the
predictors. The concept of feature hierarchies has been applied in the
machine learning domain to visual and text classification tasks, where
general features (e.g.~objects, phrases) are learned first, with
subsequent fine-tuning for lower level representations (e.g.~pixels,
words)
(\protect\hyperlink{ref-epshteinFeatureHierarchiesObject2005}{Epshtein
\& Uliman, 2005};
\protect\hyperlink{ref-girshickRichFeatureHierarchies2014}{Girshick
\emph{et al.}, 2014}). The HFR ports this concept to the linear
regression setting, where the feature hierarchy can be exploited to
increase the robustness of parameter estimates in a manner not unrelated
to its role in learning invariant representations in text and image
data. The HFR decomposes the data generating process (DGP) into a signal
graph, estimating parameters for general (shared) signal patterns
separately from the idiosyncratic contribution of each individual
predictor.

Hierarchical clustering algorithms (a sub-field of unsupervised machine
learning) present an approach to estimating the type of signal graphs
used by the HFR, and have been applied in multiple domains, including
financial time series
(\protect\hyperlink{ref-dimatteoInterestRatesCluster2004}{Di Matteo
\emph{et al.}, 2004};
\protect\hyperlink{ref-leonClusteringAlgorithmsRiskAdjusted2017}{León
\emph{et al.}, 2017};
\protect\hyperlink{ref-mantegnaHierarchicalStructureFinancial1999a}{Mantegna,
1999}; \protect\hyperlink{ref-tolaClusterAnalysisPortfolio2008}{Tola
\emph{et al.}, 2008};
\protect\hyperlink{ref-tumminelloCorrelationHierarchiesNetworks2010}{Tumminello
\emph{et al.}, 2010}). Recent applications in the financial portfolio
construction literature have resulted in an interesting conceptual
pendant to the HFR
(\protect\hyperlink{ref-lopezdepradoBuildingDiversifiedPortfolios2016}{Lopez
de Prado, 2016};
\protect\hyperlink{ref-pfitzingerConstrainedHierarchicalRisk2019}{Pfitzinger
\& Katzke, 2019};
\protect\hyperlink{ref-raffinotHierarchicalClusteringBased2016}{Raffinot,
2016}). The authors find that portfolios of financial assets can be
enhanced by replacing pairwise correlations with group-wise correlations
of asset return series. This reasoning is not unlike the mechanism by
which the HFR achieves more robust parameter estimates.

\hypertarget{the-hfr-estimator}{%
\section{\texorpdfstring{The HFR estimator
\label{method}}{The HFR estimator }}\label{the-hfr-estimator}}

\hypertarget{syntax-of-feature-hierarchies}{%
\subsection{Syntax of feature
hierarchies}\label{syntax-of-feature-hierarchies}}

Before introducing the HFR estimator, this section provides a brief
overview of the graph theoretical concepts and definitions drawn on in
the subsequent discussions.

A hierarchical representation is taken to mean the arrangement of
predictors into clusters of two or more, which are merged at nodes to
form higher levels. The predictors are the leaf nodes (i.e.~they
represent the lowest nodes in the hierarchy), while nodes at higher
levels are called internal nodes. The process of merging is repeated at
each level until all predictors are contained within a single cluster
called the root node. The node directly above any node is typically
referred to as the parent node, while the nodes below are the children.
Adjacent nodes that share a single parent are siblings. The chain of
preceding parent nodes for any node is its branch.

\begin{figure}
\centering
\begin{adjustbox}{valign=t,minipage={.49\textwidth}}
\centering
\vspace{.5cm}

\begin{center}\includegraphics{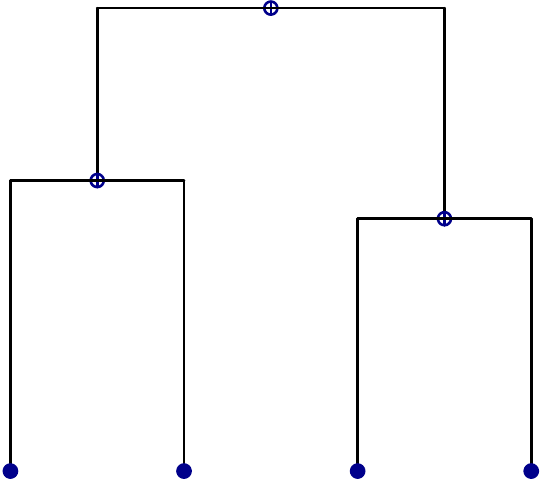} \end{center}
\end{adjustbox}
\begin{adjustbox}{valign=t,minipage={.49\textwidth}}
\centering
\begin{equation}
\mathbf{S} = \begin{bmatrix}
1 & 1 & 1 & 1 \\
1 & 1 & 0 & 0 \\
0 & 0 & 1 & 1 \\
1 & 0 & 0 & 0 \\
0 & 1 & 0 & 0 \\
0 & 0 & 1 & 0 \\
0 & 0 & 0 & 1 \\
\end{bmatrix}
\nonumber
\end{equation}
\end{adjustbox}
\caption{Example of a hierarchy dendrogram (left) and the corresponding summing matrix (right). \label{fig:dendro}}
\end{figure}

Hierarchies can be depicted graphically in dendrograms, or
mathematically in summing matrices. Figure \ref{fig:dendro} portrays a
simple hierarchy dendrogram of the illustration introduced in Section
\ref{method:framework}. There are \(K = 4\) predictors (leaf nodes), and
two subsets grouping two predictors each. The root node completes the
dendrogram.

The corresponding hierarchy summing matrix \(\mathbf{S}\) (right panel,
Fig. \ref{fig:dendro}) consists of \(D \times K\) dimensions, where
\(D = 7\) is the total number of nodes and \(K = 4\) is the number of
predictors. \(\mathbf{S}\) is invariant to the ordering of rows
(i.e.~child and parent nodes do not have to be arranged in any
particular order). However, to simplify the discussion it is presented
in a top-down order throughout this paper, starting with the root node
and ending with the leaf nodes.

Hierarchies can be cut along the \(y\)-axis of the dendrogram by drawing
a horizontal line at any height of Fig. \ref{fig:dendro}. The nodes
directly beneath the cut describe a level. In the discussions that
follow, an arbitrary level is denoted \(\ell\), and \(L\) is the total
number of levels. Fig. \ref{fig:dendro_cut} shows a cut in the
dendrogram and the summing matrix associated with that level:

\begin{figure}
\centering
\begin{adjustbox}{valign=t,minipage={.49\textwidth}}
\centering

\begin{center}\includegraphics{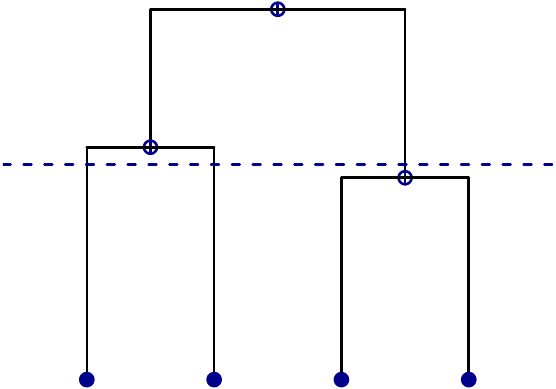} \end{center}
\end{adjustbox}
\begin{adjustbox}{valign=t,minipage={.49\textwidth}}
\centering
\vspace{.4cm}
\begin{equation}
\mathbf{S}_{\ell} = \begin{bmatrix}
1 & 0 & 0 & 0 \\
0 & 1 & 0 & 0 \\
0 & 0 & 1 & 1
\end{bmatrix}
\nonumber
\end{equation}
\end{adjustbox}
\caption{Example of a cut hierarchy dendrogram (left) and the corresponding level-specific summing matrix (right). \label{fig:dendro_cut}}
\end{figure}

A predictor hierarchy conveys information about the interrelatedness of
predictors, grouping similar predictors closely together. In the context
of the HFR, coefficients on predictors whose paths merge within the
hierarchy experience shrinkage towards a common target. The higher in
the hierarchy the merge is located, the stronger the shrinkage. In
Sections \ref{method:framework} to \ref{method:approximation}, the HFR
is introduced under the assumption of a given optimal hierarchy, while
Section \ref{method:S} introduces an algorithm to estimate
\(\mathbf{S}\).

\hypertarget{a-framework-for-group-shrinkage}{%
\subsection{\texorpdfstring{A framework for group shrinkage
\label{method:framework}}{A framework for group shrinkage }}\label{a-framework-for-group-shrinkage}}

The hierarchical feature regression is introduced using a simple
example, and following two steps: First, a decomposition of the ordinary
least squares (OLS) estimator into a sequence of node-specific estimates
in a hierarchical graph is proposed. Second, shrinkage is introduced to
the levels of the graph, resulting in the HFR estimator. The simple
example is eventually generalized in the subsequent sections.

To introduce the decomposition of the OLS estimator into hierarchical
components, take again the setting described above with \(K=4\)
standardized predictors,
\(\mathbf{x} = \{\mathbf{x}_i\}_{i = 1,...,N} \in \mathbb{R}_K\), which
are clustered into one, two and four groups, resulting in the summing
matrix in Fig. \ref{fig:dendro}, assumed to represent an optimal graph.
The matrix \(\mathbf{S}\) can be divided into sub-matrices, denoted
\(\mathbf{S}_{\ell}\), that describe the individual levels within the
feature hierarchy.

For the three levels in the example, with \(\ell = 1,2,3\), the
sub-matrices are given by \[
\mathbf{S}_1 = \begin{bmatrix} 1&1&1&1 \end{bmatrix} \;\;\;
\mathbf{S}_2 = \begin{bmatrix} 1&1&0&0 \\ 0&0&1&1 \end{bmatrix} \;\;\; 
\mathbf{S}_3 = \begin{bmatrix} 1&0&0&0 \\ 0&1&0&0 \\ 0&0&1&0 \\ 0&0&0&1 \end{bmatrix}.
\nonumber
\] Here the lowest level (\(\mathbf{S}_3\)) is an identity matrix
containing the leaf nodes.

The level-specific hierarchical features are now defined as
\(\mathbf{z}_{\ell} = \mathbf{x}\mathbf{S}_{\ell}^\top\), and the
complete hierarchical feature set is given by
\(\mathbf{z} = \mathbf{x}\mathbf{S}^\top = \begin{bmatrix} \mathbf{z}_1 & \mathbf{z}_2 & \mathbf{z}_3\end{bmatrix}\).
The hierarchical features, \(\mathbf{z}\), represent factor estimates of
the common variance contained in the child features (i.e.~the features
associated with child nodes). Under the assumption that the covariance
between the predictors' idiosyncratic components is low (such that the
mean converges to zero), the sum of the predictors represents an
estimate of the common component that is consistent up to a constant
scale. See \protect\hyperlink{ref-stockFactorModelsStructural2016}{Stock
\& Watson}
(\protect\hyperlink{ref-stockFactorModelsStructural2016}{2016b}) for a
discussion of the role of feature averaging in factor estimation.

Using the level-specific factor estimates, define \[
\mathbf{Q}_{ij} = \mathbf{z}_i^\top \mathbf{z}_j\;\;\;\text{and}\;\;\;\mathbf{Q}_{\ell y} = \mathbf{z}_{\ell}^\top\mathbf{M}_{\ell-1}y, \;\;\; i,j \in \{1,2,3\},
\] with the regression response variable
\(y = \{y_i\}_{i = 1,...,N}\in \mathbb{R}\). Here \(\mathbf{M}_{\ell}\)
is the residual maker matrix, with
\(\mathbf{M}_{\ell} = \mathbf{I}_N - \mathbf{P}_{\ell} = \mathbf{I}_N - \mathbf{z}_{\ell}\mathbf{Q}_{\ell\ell}^{-1}\mathbf{z}_{\ell}^\top\),
and \(\mathbf{M}_0 = \mathbf{I}_N\). Furthermore, \(\mathbf{I}_N\) is an
\(N\times N\) dimensional identity matrix. The role of
\(\mathbf{M}_{\ell-1}\) is to partial out the effect of each node's
branch from \(\mathbf{Q}_{\ell y}\), resulting in a regression that
updates parameter estimates using only the new information introduced at
each level. Note that in a nested hierarchical graph where each level
contains strictly more information than the preceding levels, it holds
that \begin{equation}
\mathbf{M}_{\ell-1} \equiv \prod_{i=1}^{\ell} \mathbf{M}_{\ell-i}. 
\end{equation} Thus, the information of the entire branch can be
partialled out using only \(\mathbf{M}_{\ell-1}\).

Now, with \(\boldsymbol{\hat{\beta}}_{\text{ols}}\) denoting OLS
estimates for a regression of \(y\) on \(\mathbf{x}\), a top-down
hierarchical decomposition of the OLS estimator for our problem is given
by \begin{equation}
\boldsymbol{\hat{\beta}}_{\text{ols}} = \mathbf{\hat{b}}_1 + \mathbf{\hat{b}}_2 + \mathbf{\hat{b}}_3,
\label{eq:level_specific_sum}
\end{equation} where \(\mathbf{\hat{b}}_{\ell}\) are level-specific
estimates that account for the new variation introduced at level
\(\ell\). The level-specific estimates are defined simply as the least
squares estimates for \(\mathbf{z}_{\ell}\) conditional on the path of
each node: \begin{equation}
\mathbf{\hat{b}}_{\ell} = \mathbf{S}_{\ell}^\top \mathbf{Q}_{\ell\ell}^{-1}\mathbf{Q}_{\ell y}.
\label{eq:level_specific_parameters}
\end{equation} Proposition \ref{prop:hfr} stacks the above
decomposition, and shows that the resulting estimates are numerically
equivalent to OLS estimates:

\begin{proposition}

Consider a simple regression decomposition for the case of $L = 3$ with a given summing matrix $\mathbf{S}$, hierarchical features $\mathbf{z}$ defined as above, and $\mathbf{Q}_{ij} = \mathbf{z}_i^\top\mathbf{z}_j$,
$$
\mathbf{Q}_{zz} = 
\begin{bmatrix}
\mathbf{Q}_{11} & \boldsymbol{0} & \boldsymbol{0} \\
\mathbf{Q}_{21} & \mathbf{Q}_{22} & \boldsymbol{0} \\
\mathbf{Q}_{31} & \mathbf{Q}_{32} & \mathbf{Q}_{33}
\end{bmatrix} \;\;\; \text{and} \;\;\;
\mathbf{Q}_{zy} = \mathbf{z}^\top y.
\nonumber
$$
Now the coefficient estimates $\boldsymbol{\hat{\beta}} = \mathbf{S}^\top\mathbf{Q}_{zz}^{-1} \mathbf{Q}_{zy}$ represent optimal least squares estimates of the linear slope coefficients $\boldsymbol{\beta}$ of a regression of $y$ on $\mathbf{x}$.
\label{prop:hfr}
\end{proposition}

The proof of Proposition \ref{prop:hfr} is given in \ref{appa}. Note
that \(\mathbf{Q}_{zz}\) can be written as
\(\mathbf{Q}_{zz} = (\mathbf{z}^\top\mathbf{z})\odot \mathbf{H}\), where
\(\odot\) is the element-wise multiplication operator, and
\(\mathbf{H}\) is a matrix of ones with the block-wise upper triangle
set to zero: \[
\mathbf{H} = 
\begin{bmatrix}
\boldsymbol{1} & \boldsymbol{0} & \boldsymbol{0} \\
\boldsymbol{1} & \boldsymbol{1} & \boldsymbol{0} \\
\boldsymbol{1} & \boldsymbol{1} & \boldsymbol{1}
\end{bmatrix}, \;\;\; \boldsymbol{1} = 
\begin{bmatrix}
1 & \cdots & 1 \\
\vdots & \ddots & \vdots \\
1 & \cdots & 1
\end{bmatrix}, \;\;\; \boldsymbol{0} = 
\begin{bmatrix}
0 & \cdots & 0 \\
\vdots & \ddots & \vdots \\
0 & \cdots & 0
\end{bmatrix}.
\] The matrix \(\mathbf{H}\) eliminates bottom-up conditional effects
from the precision matrix, which are represented by the upper
block-triangular entries. Conversely, the lower block-triangular entries
represent conditional effects flowing down the hierarchy from the root
node towards the leaf nodes (i.e.~top-down effects).

As shown in \ref{appa}, Proposition \ref{prop:hfr} is equivalent to the
chain of level-specific estimates introduced in Eq.
\ref{eq:level_specific_sum}, with \begin{equation}
\mathbf{S}^\top\mathbf{Q}_{zz}^{-1} \mathbf{Q}_{zy}  = \mathbf{\hat{b}}_1 + \mathbf{\hat{b}}_2 + \mathbf{\hat{b}}_3. 
\label{eq:hierarchical_decomposition}
\end{equation} In sum, therefore, the hierarchical decomposition
consists of an additive chain of level-specific estimates (Eq.
\ref{eq:level_specific_sum}) that iteratively adjust for idiosyncratic
variation in the fitting process in a top-down manner (Eq.
\ref{eq:level_specific_parameters}), until at the final level
(\(\ell = 3\) in the example) all explainable variation is accounted
for.

Fig. \ref{fig:level_specific_sum} plots a dendrogram of the
decomposition, expanding the root node such that each level is
represented by a band of unit width. As shown later in the section, the
width of each level-specific band will come to represent the proportion
to which information introduced at that level is incorporated into the
HFR estimates. Each \(\mathbf{\hat{b}}_{\ell}\) adjusts the coefficients
based on the new cluster information at \(\ell\), with a single cluster
at \(\mathbf{\hat{b}}_{1}\), two clusters at \(\mathbf{\hat{b}}_{2}\)
and four clusters at \(\mathbf{\hat{b}}_{3}\):

\begin{figure}[H]

{\centering \includegraphics{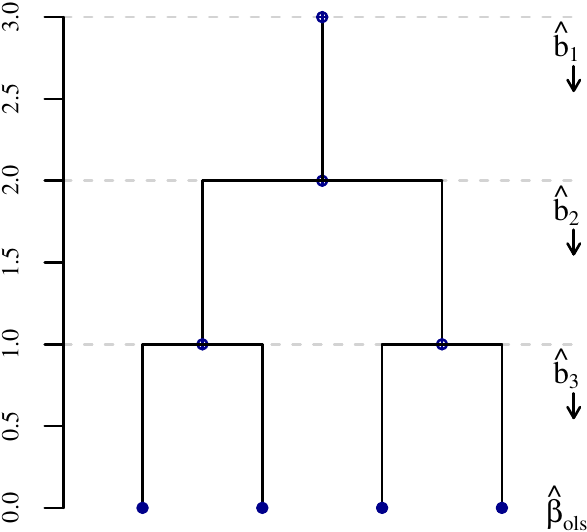} 

}

\caption{Dendrogram of the level-wise decomposition of the OLS estimator.\label{fig:level_specific_sum}}\label{fig:unnamed-chunk-3}
\end{figure}

While this decomposition seems trivial at first glance, it can be used
as the basis for a regularized regression. The HFR estimator shrinks the
extent to which levels are permitted to adjust for new variation,
resulting in estimates that are biased towards higher-level
representations in the form of group targets for clusters of predictors,
with lower levels not permitted to adjust fully to the variation
contained in them.

In the simplest form, one could add a shrinkage coefficient to Eq.
\ref{eq:level_specific_sum}, such that \begin{equation}
\boldsymbol{\hat{\beta}}_{\text{hfr}} = \sum_{\ell = 1}^3 \theta_{\ell} \mathbf{\hat{b}}_{\ell},
\label{eq:chain_w_shrinkage}
\end{equation} where \(\theta_{\ell}\) is the \(\ell\)th shrinkage
coefficient, with \(0 \leq\theta_{\ell}\leq \theta_{\ell-1}\) and
\(0\leq\theta_1\leq 1\). For instance, if \(\theta_2 = \theta_3 = 0\)
and \(\theta_1 = 1\), the estimates are reduced to
\(\boldsymbol{\hat{\beta}}_{\text{hfr}} = \mathbf{\hat{b}}_1\), which is
equivalent to a single group mean across all parameters. The
monotonicity constraint on \(\theta_{\ell}\) ensures that --- given that
the hierarchy represents a nested information set --- information that
is removed at one level is not subsequently reintroduced at a lower
level.

Fig. \ref{fig:chain_w_shrinkage} plots two shrunken dendrograms with the
degree of shrinkage represented by the distance between two levels and
equal to \(\theta_{\ell}\). The left panel of Fig.
\ref{fig:chain_w_shrinkage} represents moderate shrinkage, while the
right panel removes an entire level:

\begin{figure}[H]

{\centering \includegraphics{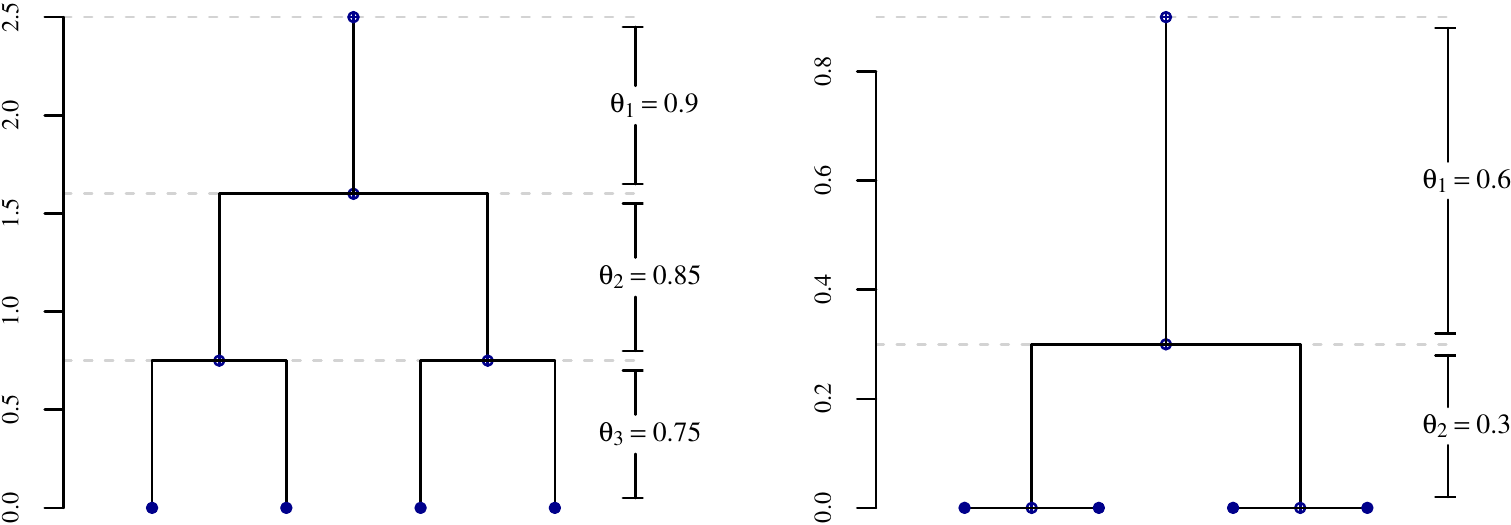} 

}

\caption{Dendrogram of the level-wise decomposition of the OLS estimator with shrinkage represented by the distance between levels.\label{fig:chain_w_shrinkage}}\label{fig:unnamed-chunk-4}
\end{figure}

In the stacked form of Proposition \ref{prop:hfr}, Eq.
\ref{eq:chain_w_shrinkage} introduces a shrinkage matrix, such that
\begin{equation}
\boldsymbol{\hat{\beta}}_{\text{hfr}} = 
\mathbf{S}^\top(\mathbf{Q}_{zz} \odot \boldsymbol{\Theta})^{-1} \mathbf{Q}_{zy}
\label{eq:shrink_mult}
= \mathbf{S}^\top\left( \begin{bmatrix}
\mathbf{Q}_{11} & \boldsymbol{0} & \boldsymbol{0} \\
\mathbf{Q}_{21} & \mathbf{Q}_{22} & \boldsymbol{0} \\
\mathbf{Q}_{31} & \mathbf{Q}_{32} & \mathbf{Q}_{33}
\end{bmatrix} \odot \begin{bmatrix}
\boldsymbol{\Theta}_1 & \boldsymbol{\Theta}_2 & \boldsymbol{\Theta}_3
\end{bmatrix}  \right)^{-1}
\mathbf{Q}_{zy}.
\end{equation} Here \(\boldsymbol{\Theta}_{\ell}\) governs the extent of
shrinkage for level \(\ell\), with \[
\boldsymbol{\Theta}_{\ell} = \begin{bmatrix} 
\theta_{\ell}^{-1} & \cdots & \theta_{\ell}^{-1} \\
\vdots & \ddots & \vdots \\
\theta_{\ell}^{-1} & \cdots & \theta_{\ell}^{-1} 
\end{bmatrix}.
\] Eq. \ref{eq:shrink_mult} is the HFR estimator under the assumption
that the hierarchy (\(\mathbf{S}\)) as well as the extent of shrinkage
(\(\boldsymbol{\Theta}\)) are given. Since the framework permits the
exclusion of entire levels from the regression (by setting
\(\theta_{\ell} = 0\)), it can be used as a tool to select a
parsimonious hierarchy based on a (potentially large) set of input
levels. As shown in subsequent sections, this property will be useful
for the estimation of \(\mathbf{S}\).

The framework described by Eq. \ref{eq:shrink_mult} can become
arbitrarily complex, including a large number of levels that permit a
high degree of nuance with respect to the nature and strength of
regularization. At its core, however, it remains a decomposition of each
parameter into a chain of parameters, which captures successively more
idiosyncratic signal, and which is subsequently regularized, resulting
in overall shrinkage towards a more general and less idiosyncratic
representation of the data generating process. A key ingredient for this
form of group shrinkage is determining the optimal extent of shrinkage
for each hierarchical level (i.e.~for the elements of the parameter
chain). Section \ref{method:hfr} discusses an appropriate loss function
that can be used to obtain optimal shrinkage coefficients.

\hypertarget{optimal-shrinkage}{%
\subsection{\texorpdfstring{Optimal shrinkage
\label{method:hfr}}{Optimal shrinkage }}\label{optimal-shrinkage}}

Generalizing the definition in Eq. \ref{eq:shrink_mult} to an arbitrary
hierarchy, the hierarchical feature regression estimates,
\(\boldsymbol{\hat{\beta}}_{\text{hfr}}\), are given by \begin{equation}
\boldsymbol{\hat{\beta}}_{\text{hfr}} = \mathbf{S}^\top(\mathbf{z}^\top\mathbf{z} \odot \mathbf{H} \odot  \boldsymbol{\Theta})^{-1}\mathbf{z}^\top y,
\label{eq:hfr}
\end{equation} where \(\mathbf{H}\), as before, is a matrix of ones and
zeros, which ensures that the block-wise upper triangle of
\(\mathbf{z}^\top\mathbf{z}\) is zero. \(\boldsymbol{\Theta}\) is a
\(D\times D\) matrix controlling the extent of shrinkage on a
level-by-level basis. Letting \(\boldsymbol{\Theta}_{\ell}\) contain
that subset of columns of \(\boldsymbol{\Theta}\) associated with nodes
in level \(\ell\), \(\boldsymbol{\Theta}\) is defined as \[
\boldsymbol{\Theta} = \begin{bmatrix} \boldsymbol{\Theta}_1 & \cdots & \boldsymbol{\Theta}_L \end{bmatrix}
\;\;\; \text{where} \;\;\;
\boldsymbol{\Theta}_{\ell} = 
\begin{bmatrix} 
\theta_{\ell}^{-1} & \cdots & \theta_{\ell}^{-1} \\
\vdots & \ddots & \vdots \\
\theta_{\ell}^{-1} & \cdots & \theta_{\ell}^{-1} 
\end{bmatrix} \;\;\; \forall \;\;\; \ell \in 1,...,L.
\] The extent of shrinkage is therefore governed entirely by the
\(L \times 1\) vector of level-specific shrinkage coefficients
\(\boldsymbol{\theta} = \begin{pmatrix} \theta_1 & \cdots & \theta_L \end{pmatrix}\).
When \(\boldsymbol{\theta} = \boldsymbol{1}\) there is no shrinkage,
leading to the OLS solution as shown in Proposition \ref{prop:hfr}. If
any \(\theta_{\ell} < 1\), the parameters associated with level \(\ell\)
are regularized, where \(\theta_{\ell} \rightarrow 0\) constitutes
maximum shrinkage. Note that when \(\theta_{1}<1\), the entire
parameter-norm is shrunken.

When \(\mathbf{S}\) includes the maximum possible number of levels, with
\(L = K\) so that each level comprises exactly one more cluster than the
preceding level, the shrinkage vector \(\boldsymbol{\theta}\) has the
useful property that its sum (i.e.~the sum of all level-specific
adjustments to new variation) is equal to the effective model size, as
captured by the effective degrees of freedom (\(\nu_{\text{eff}}\)):

\begin{proposition}

With an HFR projection matrix given by $\mathbf{P}_{\textup{hfr}} = \mathbf{z} (\mathbf{z}^\top\mathbf{z} \odot \mathbf{H} \odot \boldsymbol{\Theta})^{-1}\mathbf{z}^\top$, and the effective model degrees of freedom defined in the usual manner using the trace of the projection matrix, $\nu_{\textup{eff}} = \textup{tr}(\mathbf{P}_{\textup{hfr}})$, it holds that
\begin{equation}
\nu_{\textup{eff}} \equiv \sum_{\ell=1}^L \theta_{\ell}, \nonumber
\end{equation}
when $L = K$ distinct levels are included in the hierarchy described by $\mathbf{S}$.
\label{prop:shrinkage_vector}
\end{proposition}

The proof of Proposition \ref{prop:shrinkage_vector} is given in
\ref{appb}.

With this definition in hand, an information theoretically motivated
approach to the determination of an optimal shrinkage vector,
\(\boldsymbol{\theta}^*\), is to impose a constraint on the effective
model size. Defining a hyperparameter, \(\kappa\), that represents the
effective model size (normalized to a value between 0 and 1), the
optimal shrinkage vector is the solution that maximizes fit subject to
the constraint \begin{equation}
\sum_{\ell = 1}^L\theta_{\ell} = \kappa K.
\end{equation} When \(\kappa = 1\), the problem is unconstrained, with
\(\nu_{\text{eff}}=K\) and
\(\boldsymbol{\hat{\beta}}_{\text{hfr}} = \boldsymbol{\hat{\beta}}_{\text{ols}}\).
Conversely, when \(\kappa < 1\), the model fit is maximized given a
predetermined value for \(\nu_{\text{eff}}\). Expressing the
optimization in terms of the HFR loss, the optimal extent of shrinkage
conditional on hyperparameter \(\kappa\), is given by \begin{align}
\boldsymbol{\theta}^*_{\kappa} = \arg \min_{\boldsymbol{\theta}} & \left[ N^{-1} (\mathbf{x}\boldsymbol{\hat{\beta}}_{\text{hfr}} - y)^\top(\mathbf{x}\boldsymbol{\hat{\beta}}_{\text{hfr}} - y) \right] \label{eq:hfr_exact_loss} \\
\text{s.t.}\;\;\;&0\leq\theta_{\ell}\leq\theta_{\ell-1} \; \forall\; \ell>1, \nonumber \\
&0\leq\theta_1\leq 1, \nonumber \\
&\sum_{\ell=1}^L \theta_{\ell} = \kappa K. \nonumber
\end{align} Eq. \ref{eq:hfr_exact_loss} trades off goodness-of-fit
against parsimony, where the hyperparameter \(\kappa\) tilts the global
trade-off towards goodness-of-fit as \(\kappa \rightarrow 1\), or
parsimony as \(\kappa \rightarrow 0\). Fig.
\ref{fig:level_specific_sum_full} plots the complete dendrogram for the
example problem, with \(L = 4\) levels, where \(\ell=3\) had been
omitted previously for the sake of simplicity. The total height of the
dendrogram is now exactly equal to the effective model size
(\(\kappa=1\)). In fact, the definition of \(\kappa\) ensures that the
hyperparameter represents the overall size of the optimal HFR graph as a
percentage of \(K\), with a shallower hierarchy as
\(\kappa \rightarrow 0\):

\begin{figure}[H]

{\centering \includegraphics{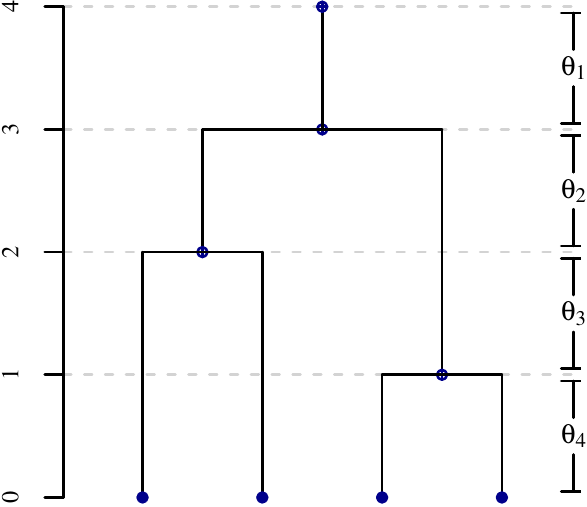} 

}

\caption{Dendrogram of the level-wise decomposition of the OLS estimator with $L = K$ levels.\label{fig:level_specific_sum_full}}\label{fig:unnamed-chunk-5}
\end{figure}

The following section demonstrates how the optimal shrinkage vector
\(\boldsymbol{\theta}_{\kappa}^*\) can be obtained efficiently for any
given value of \(\kappa\) using quadratic programming to solve Eq.
\ref{eq:hfr_exact_loss}.

\hypertarget{recasting-the-hfr-as-a-model-average}{%
\subsection{\texorpdfstring{Recasting the HFR as a model average
\label{method:approximation}}{Recasting the HFR as a model average }}\label{recasting-the-hfr-as-a-model-average}}

The HFR estimates in Eq. \ref{eq:hfr} can be restated as the dot product
of level-specific estimates and a transformed shrinkage vector, such
that \begin{equation}
\boldsymbol{\hat{\beta}}_{\text{hfr}} = \boldsymbol{\hat{\mathcal{B}}}\boldsymbol{\phi}.
\label{eq:hfr_phi}
\end{equation} Here \(\boldsymbol{\hat{\mathcal{B}}}\) stacks
unconditional level-specific estimates (unconditional with respect to
preceding levels in the hierarchy), such that with
\(\boldsymbol{\hat{\mathcal{B}}} = \begin{bmatrix} \mathbf{\hat{w}}_1 & \cdots & \mathbf{\hat{w}}_L\end{bmatrix}\),
\begin{equation}
\mathbf{\hat{w}}_{\ell} = \mathbf{S}_{\ell}^\top\left(\mathbf{z}_{\ell}^\top\mathbf{z}_{\ell}\right)^{-1}\mathbf{z}_{\ell}^\top y.
\end{equation} Note that \(\mathbf{\hat{w}}_{\ell}\) is an unconditional
counterpart to \(\mathbf{\hat{b}}_{\ell}\), where the effect of each
node's branch has not been partialled out. Furthermore,
\(\boldsymbol{\phi}\) is a transformation of \(\boldsymbol{\theta}\)
that satisfies the equality
\(\boldsymbol{\theta}=\boldsymbol{\omega}^\top\boldsymbol{\phi}\), where
\(\boldsymbol{\omega}\) is a lower triangular matrix, resulting in \[
\boldsymbol{\phi} = \begin{cases}
\theta_{\ell} - \theta_{\ell+1} &\text{when }\ell < L\\
\theta_{\ell} &\text{otherwise.}
\end{cases}
\] The derivation of Eq. \ref{eq:hfr_phi} is given in \ref{appc}, and
follows directly from the introduction of shrinkage weights to the
calculations in \ref{appa}. By reformulating the problem in an
unconditional manner, \(\mathbf{\hat{w}}_{\ell}\) can be computed in
parallel for each level, and the optimization of \(\boldsymbol{\theta}\)
can be split into two consecutive steps: (i) estimating level-specific
regressions (\(\boldsymbol{\hat{\mathcal{B}}}\)), and (ii) constructing
the optimal shrinkage hierarchy by optimizing \(\boldsymbol{\phi}\).

Eq. \ref{eq:hfr_phi} resembles a model-averaging estimator, where the
models \(\boldsymbol{\hat{\mathcal{B}}}\) are averaged by the weights
\(\boldsymbol{\phi}\). Mallows model averaging (MMA), for instance,
represents a close mathematical pendant, where the weighting vector is
obtained by minimizing the Mallows information criterion
(\protect\hyperlink{ref-hansenLeastSquaresModel2007}{Hansen, 2007};
\protect\hyperlink{ref-mallowsCommentsCP1973}{Mallows, 1973}). The
optimal shrinkage problem of the HFR can correspondingly be thought of
as the minimization of a custom information criterion (Eq.
\ref{eq:hfr_exact_loss}) to determine the optimal vector
\(\boldsymbol{\phi}^*_{\kappa}\) (which, by extension, yields
\(\boldsymbol{\theta}^*_{\kappa}\)). Importantly, the information
theoretic model-averaging problem is quadratic in its weights
(i.e.~quadratic in \(\boldsymbol{\phi}\)), and can be solved
analytically using quadratic programming algorithms.

Following this reasoning, the level-specific coefficients in
\(\boldsymbol{\hat{\mathcal{B}}}\) are used to reformulate the
optimization, such that with
\(\mathbf{\hat{y}} = \mathbf{x}\boldsymbol{\hat{\mathcal{B}}}\),
\(\mathbf{U} = \frac{1}{N}\mathbf{\hat{y}}^\top\mathbf{\hat{y}}\) and
\(\mathbf{V} = \frac{2}{N}\mathbf{\hat{y}}^\top y\) \begin{align}
\boldsymbol{\phi}_{\kappa}^* = \arg \min_{\boldsymbol{\phi}} & \left[ \boldsymbol{\phi}^\top\mathbf{U}\boldsymbol{\phi} - \mathbf{V}^\top\boldsymbol{\phi} \right] \label{eq:qp} \\
\text{s.t.} \;\;\; &\boldsymbol{0}\leq \boldsymbol{\phi}\leq \boldsymbol{1} \nonumber \\
&\boldsymbol{\phi}^\top\boldsymbol{\omega}\mathbf{1} = \kappa K. \nonumber
\end{align} Since the original shrinkage vector can be expressed as
\(\boldsymbol{\theta} = \boldsymbol{\omega}^\top \boldsymbol{\phi}\),
the constraints in Eq. \ref{eq:qp} are identical to the constraints in
Eq. \ref{eq:hfr_exact_loss}, and
\(\boldsymbol{\theta}_{\kappa}^* = \boldsymbol{\omega}^\top\boldsymbol{\phi}_{\kappa}^*\).
Note that the monotonicity constraint collapses to a simple weight
constraint on \(\boldsymbol{\phi}\).

The HFR estimates given by
\(\boldsymbol{\hat{\beta}}_{\text{hfr}} = \boldsymbol{\hat{\mathcal{B}}}\boldsymbol{\phi}_{\kappa}^*\)
have thus far assumed a given hierarchy, encoded in \(\mathbf{S}\). The
aim of the HFR is to estimate \(\mathbf{S}\) in a supervised manner,
which conceptually requires selecting the composition of predictor
groups at each level that minimizes Eq. \ref{eq:hfr_exact_loss}. This is
a computationally intractable combinatorial problem. Instead, the
following section suggests a feasible and computationally efficient
algorithm for arriving at a graph estimate based on the similarity of
the predictors' explanatory structure in \(y\), using supervised
hierarchical clustering.

\hypertarget{graph-estimation}{%
\subsection{\texorpdfstring{Graph estimation
\label{method:S}}{Graph estimation }}\label{graph-estimation}}

The graph-based decomposition of linear regression parameters introduced
in the preceding sections assumes a hierarchical arrangement of
predictors into \(L = K\) levels that are captured in \(\mathbf{S}\).
Here \(\mathbf{S}\) contains the maximum number of levels possible in a
nested hierarchical tree, while \(\boldsymbol{\theta}^*_{\kappa}\)
selects a parsimonious hierarchy by reducing the weight of individual
levels, or removing levels from the hierarchy entirely. In order to
estimate \(\mathbf{S}\), I propose a supervised hierarchical clustering
algorithm, that merges variables based on the similarity of their
explanatory component with respect to \(y\).

A typical approach to (unsupervised) hierarchical clustering constructs
a dissimilarity matrix \(\boldsymbol{\mathcal{D}}\) that encodes
information about the predictor set (e.g.~the (inverse) pairwise
correlation coefficients, or distances), and recursively merges the
predictors or clusters with the lowest cluster distance
(\protect\hyperlink{ref-maimonDataMiningKnowledge2010}{Maimon \& Rokach,
2010}). The aim of a supervised rendition of a hierarchical clustering
algorithm is to merge those clusters that maximize the goodness-of-fit
of a regression of \(y\) on the appropriate cluster features
\(\mathbf{z}_{\ell}\) at each \(\ell\).\footnote{This differs
  conceptually from a traditional understanding of supervised
  clustering, where true cluster labels are used to train a model, with
  the aim of predicting new cluster labels.} Two predictors or clusters
are deemed similar, if merging them leads to a comparatively small
increase in the regression error, or conversely, a comparatively small
decline in the goodness-of-fit.

Consider the previous example of a regression of \(y\) on four
predictors \(\mathbf{x}\), with the estimated regression fit given by:
\begin{equation}
\hat{y} = \hat{\beta}_1x_1 + \hat{\beta}_2x_2 + \hat{\beta}_3x_3 + \hat{\beta}_4x_4.
\end{equation} A merge of any two predictors \(i,j\) results in:
\begin{equation}
\hat{y} = \hat{\beta}_{ij}(x_i + x_j) + \mathbf{x}_{-ij}\boldsymbol{\hat{\beta}}_{-ij},\label{eq:reg_merge}
\end{equation} where \(\mathbf{x}_{-ij}\) contains all remaining
predictors.

A merge is therefore akin to the imposition of an equality constraint on
the associated coefficients \(\beta_i\) and \(\beta_j\). This equality
constraint is least costly (in terms of goodness-of-fit), when the
conditional effect of \(x_i\) and \(x_j\) on \(y\) is similar. That is
when \begin{equation}
r_{x_i,y|\mathbf{x}_{-ij}}\approx r_{x_j,y|\mathbf{x}_{-ij}}. \label{eq:reg_partial}
\end{equation} Here, \(r_{x_i,y|\mathbf{x}_{-ij}}\) is the partial
correlation between \(x_i\) and \(y\) conditional on
\(\mathbf{x}_{-ij}\).

An intuitively appealing and computationally feasible alternative to the
estimation of regression fits for each possible cluster combination (Eq.
\ref{eq:reg_merge}), is therefore to examine the similarity of the
partial correlation coefficients. If the within-cluster variance of
partial correlations is small (Eq. \ref{eq:reg_partial}), the cost of
the equality constraint, and, by extension, the reduction in
goodness-of-fit, can be expected to be low.

\protect\hyperlink{ref-wardHierarchical1963}{Ward}
(\protect\hyperlink{ref-wardHierarchical1963}{1963}) outlines an
agglomerative clustering algorithm that achieves just this: merging
clusters based on the minimum additional within-cluster variance
introduced by the merge. The author shows that the approach can be
reduced to a clustering based on the Euclidean distances between the
input vectors. The algorithm begins by placing each row in
\(\boldsymbol{\mathcal{D}}\) into a cluster of its own, and iteratively
merges those clusters that result in the minimum increase in overall
within-cluster variance. Clusters are merged a total of \(K-1\) times,
until all rows in \(\boldsymbol{\mathcal{D}}\) are contained in a single
cluster, and \(L = K\) levels have been formed. A detailed description
of \protect\hyperlink{ref-wardHierarchical1963}{Ward}
(\protect\hyperlink{ref-wardHierarchical1963}{1963}) clustering can be
found in \protect\hyperlink{ref-kaufmanFindingGroupsData2005}{Kaufman \&
Rousseeuw} (\protect\hyperlink{ref-kaufmanFindingGroupsData2005}{2005})
and \protect\hyperlink{ref-everitt_cluster_2011}{Everitt \emph{et al.}}
(\protect\hyperlink{ref-everitt_cluster_2011}{2011}). The algorithm is
implemented using the \texttt{cluster} package in the statistical
computing language \texttt{R}
(\protect\hyperlink{ref-maechlerClusterClusterAnalysis2019}{Maechler
\emph{et al.}, 2019};
\protect\hyperlink{ref-rcoreteamLanguageEnvironmentStatistical2018}{R
Core Team, 2018}).

Substituting partial correlations for \(\boldsymbol{\mathcal{D}}\)
results in a supervised hierarchical clustering algorithm. However,
since conditioning on \(\mathbf{x}_{-ij}\) is at best imprecise and at
worst unfeasible in the high-dimensional setting, an approximate
supervised dissimilarity matrix can instead be defined based on
bivariate partial correlations, such that: \begin{equation}
\boldsymbol{\mathcal{D}}_{y} = \{r_{x_i,y | x_j}\}_{i,j = 1,...,K}, \;\;\; \text{and} \;\;\; r_{x_i,y | x_j} = \frac{r_{y,x_i} - r_{y,x_j} r_{x_i,x_j}}{\sqrt{(1 - r_{y,x_j}^2)(1 - r_{x_i,x_j}^2)}},  \;\;\; i\neq j.
\label{eq:dist}
\end{equation} Note that \(\text{diag}(\boldsymbol{\mathcal{D}}_{y})\)
is undefined so that, letting \(\boldsymbol{d}_i\) denote the \(i\)th
row, \(||\boldsymbol{d}_i-\boldsymbol{d}_j||\) measures the distance
between \(\{r_{x_i,y|x_{k}}\}_{k\notin i,j}\) and
\(\{r_{x_j,y|x_{k}}\}_{k\notin i,j}\) (i.e.~the distance between the
bivariate partial correlations conditioning on all predictors in
\(\mathbf{x}_{-ij}\) individually).

The matrix \(\boldsymbol{\mathcal{D}}_y\) results in a sign-sensitive
clustering of parameters (positive and negative coefficients tend to be
clustered separately). However, at the highest levels in the hierarchy,
clusters will invariably contain effects with mixed signs. To ensure
sign-invariance, with shrinkage towards absolute group targets, the
summing matrix \(\mathbf{S}\) must be adjusted such that
\begin{equation}
\mathbf{S}_{i} = \mathbf{S}_{i}^{+} \odot \text{sign}(\boldsymbol{\rho}^\top\mathbf{S}_i^+ - \mathbf{1}),
\label{eq:sign_adjustment}
\end{equation} where \(\mathbf{S}_i\) is a row in in \(\mathbf{S}\), and
\(\boldsymbol{\rho} = \text{cor}(\mathbf{x})\). The matrix
\(\mathbf{S}^{+}\) is the unadjusted (positive-only) summing matrix.
This ensures that when coefficients with opposite signs are contained in
a single cluster, their effect is mirrored and not averaged.\footnote{As
  an aside, the HFR can be made entirely sign-invariant, permitting
  negatively correlated predictors with a similar explanatory effect on
  \(y\) --- albeit with opposite signs --- to be clustered adjacently.
  This is achieved by using the absolute partial correlation matrix,
  \(|\boldsymbol{\mathcal{D}}_{y}|\). Such an approach is useful when
  the sign is not deemed to convey meaningful information, with only the
  absolute size of the coefficients being relevant.}

The combination of \protect\hyperlink{ref-wardHierarchical1963}{Ward}
(\protect\hyperlink{ref-wardHierarchical1963}{1963}) clustering and
partial correlations between \(y\) and \(\mathbf{x}\) produces a
supervised hierarchical clustering algorithm that merges clusters based
on the within-cluster variance of the partial correlations --- a method
that is analogous to the minimization of the cost of the hierarchical
constraint encoded in \(\mathbf{S}\). Since the hierarchical constraint
increases the regression error at each merge, its minimization is
analogous to a selection of cluster-splits using a goodness-of-fit
criterion, but can be implemented within the efficient framework of
agglomerative clustering algorithms.

\hypertarget{deterministic-terms-standard-errors-and-further-issues}{%
\subsection{\texorpdfstring{Deterministic terms, standard errors and
further issues
\label{method:constant}}{Deterministic terms, standard errors and further issues }}\label{deterministic-terms-standard-errors-and-further-issues}}

The preceding discussions have abstracted from deterministic elements in
the regression. Including these is exceedingly simple, and can be
achieved by adjusting the level-specific regressions in
\(\boldsymbol{\hat{\mathcal{B}}}\). Letting
\(\mathbf{M} = \{\mathbf{M}_i\}_{i = 1,...,N} \in \mathbb{R}_M\) be a
matrix of \(M\) deterministic elements (e.g.~a vector of ones), with the
associated parameter estimates \(\mathbf{\hat{m}}\), the level-specific
regression becomes: \begin{equation}
\begin{bmatrix}
\mathbf{\hat{m}}_{\ell} \\
\mathbf{\hat{w}}_{\ell}
\end{bmatrix} = \mathbf{\tilde{S}}_{\ell}^\top(\mathbf{\tilde{z}}_{\ell}^\top\mathbf{\tilde{z}}_{\ell} )^{-1}\mathbf{\tilde{z}}_{\ell}^\top y,
\label{eq:hfr_ell_const}
\end{equation} where
\(\mathbf{\tilde{z}}_{\ell} = \begin{bmatrix} \mathbf{M} & \mathbf{z}_{\ell} \end{bmatrix}\),
and \(\mathbf{\tilde{S}_{\ell}}\) expands \(\mathbf{S}_{\ell}\) such
that \[
\mathbf{\tilde{S}}_{\ell} = \begin{bmatrix}
\mathbf{I}_M & \boldsymbol{0} \\
\boldsymbol{0} & \mathbf{S}_{\ell}
\end{bmatrix}.
\] Since deterministic elements are exogenous to the estimation of the
hierarchy, the corresponding parameters are not regularized. Apart from
a regression constant, deterministic elements can include statistical
features such as trends or dummy variables, or simply predictors that,
for one reason or another, are better represented outside of the feature
hierarchy. All applications in this paper contain a deterministic
element in the form of a regression constant.

The analogy of the HFR to a model average over level-specific
regressions can furthermore be extended to obtain approximate standard
errors of the parameter estimates. Since level-specific standard errors,
\(\hat{\text{se}}(\mathbf{\hat{w}}_{\ell})\), are readily retrieved from
the level-specific regressions, the average standard errors
\(\hat{\text{se}}(\boldsymbol{\hat{\beta}}_{\text{hfr}})\) can be
obtained following \protect\hyperlink{ref-burnhamMultimodel2004}{Burnham
\& Anderson} (\protect\hyperlink{ref-burnhamMultimodel2004}{2004}), with
\begin{equation}
\hat{\text{se}}(\boldsymbol{\hat{\beta}}_{\text{hfr}}) = \sum_{\ell = 1}^L \phi_{\ell}\sqrt{\hat{\text{se}}(\mathbf{\hat{w}}_{\ell})^2 + (\mathbf{\hat{w}}_{\ell} - \mathbf{\hat{\bar{w}}}_{\boldsymbol{\phi}})^2},
\end{equation} where the weighted average parameters
\(\mathbf{\hat{\bar{w}}}_{\boldsymbol{\phi}}\) are simply the HFR
estimates \(\boldsymbol{\hat{\beta}}_{\text{hfr}}\). It is important to
note that for purposes of inference
\(\hat{\text{se}}(\boldsymbol{\hat{\beta}}_{\text{hfr}})\) are
understated. For instance, the graph estimation error embedded in
\(\mathbf{S}\) is omitted entirely. Nonetheless, the standard errors
provide valuable information about the average significance along the
branch of each variable in the hierarchy, and can be useful to prune
noise clusters and to inform sparse model selection, as illustrated in
the following section. Once again, in the absence of shrinkage, with
\(\boldsymbol{\theta}=\mathbf{1}\), the standard errors
\(\hat{\text{se}}(\boldsymbol{\hat{\beta}}_{\text{hfr}})\) are
equivalent to the standard errors of the OLS regression.

An additional tool in understanding the role of the optimal parameter
graph is to examine the level-wise decomposition of the coefficient of
determination. Letting the model fit up to the \(\ell\)th level be given
by \begin{equation}
\hat{y}_{\rightarrow\ell} = \sum_{i = 0}^{\ell-1} \theta_{\ell-i} \mathbf{x}\mathbf{\hat{b}}_{\ell-i},
\end{equation} the cumulative coefficient of determination can be
defined in the usual manner, with \begin{equation}
R^2_{\rightarrow\ell} = 1 - \frac{\sum_{i = 1}^N([\hat{y}_{\rightarrow\ell}]_i - y_i)^2}{\sum_{i = 1}^N(y_i - \bar{y})^2}.
\label{eq:levelR2}
\end{equation} When \(\ell=L\), this simply results in the total \(R^2\)
of the HFR fit. However, the level-wise formulation in Eq.
\ref{eq:levelR2} also yields contributions of each individual level to
the overall coefficient of determination, where
\(R^2_{\ell} = R^2_{\rightarrow\ell} - R^2_{\rightarrow\ell-1}\) and
\(\sum_{\ell}R^2_{\ell} = R^2\). In the plots in Section
\ref{empirical}, the level contributions are added to the dendrograms as
bars with darker colors suggesting a larger contribution of that level,
as illustrated in Fig. \ref{fig:level_specific_sum_bar}:

\begin{figure}[H]

{\centering \includegraphics{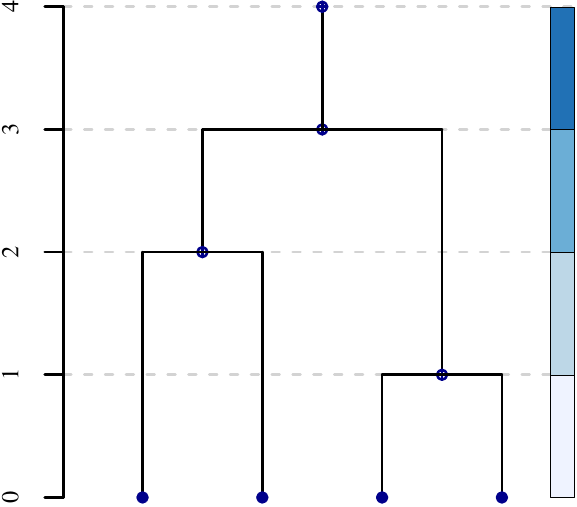} 

}

\caption{Illustrative dendrogram of the level-wise decomposition of the HFR estimates with level-specific contributions on the right.\label{fig:level_specific_sum_bar}}\label{fig:unnamed-chunk-6}
\end{figure}

As a final issue, the discussion has thus far assumed \(K < N-M\). When
\(K \geq N-M\), the level-specific regressions for all \(\ell \geq N-M\)
cannot be computed. Since the lowest levels group predictors with the
highest similarity, the simplest remedy is to prune all levels where
\(\ell \geq N-M\). This leaves a total of \(L = N-M-1\) levels with no
effect on the structure of the HFR, with the sole exception that the
constraint in Eq. \ref{eq:qp} substitutes \(\kappa(N-M-1)\) for
\(\kappa K\): \begin{align}
\boldsymbol{\phi}_{\kappa}^* = \arg \min_{\boldsymbol{\phi}} & \left[ \boldsymbol{\phi}^\top\mathbf{U}\boldsymbol{\phi} - \mathbf{V}^\top\boldsymbol{\phi} \right] \\
\text{s.t.} \;\;\; &\boldsymbol{0}\leq \boldsymbol{\phi}\leq \boldsymbol{1} \nonumber \\
&\boldsymbol{\phi}^\top\boldsymbol{\omega}\mathbf{1} = \kappa (N-M-1). \nonumber
\end{align} An implementation of the HFR algorithm and the issues
discussed in this paper is provided in the \texttt{hfr} package
available on the Comprehensive \texttt{R} Archive Network (CRAN) for the
statistical computing language \texttt{R}
(\protect\hyperlink{ref-pfitzingerHFR2022}{Pfitzinger, 2022}).

\hypertarget{a-case-study-determinants-of-economic-growth}{%
\section{\texorpdfstring{A case study: Determinants of economic growth
\label{empirical}}{A case study: Determinants of economic growth }}\label{a-case-study-determinants-of-economic-growth}}

The HFR is useful both as a regression estimator and as a tool to garner
insights into the effect structure underlying the estimated statistical
model. In this section, I propose an analysis workflow that uses the HFR
to understand an empirical problem and to obtain robust out-of-sample
predictions. The data is taken from
\protect\hyperlink{ref-sala-i-martinDeterminantsLongTermGrowth2004}{Sala-I-Martin
\emph{et al.}}
(\protect\hyperlink{ref-sala-i-martinDeterminantsLongTermGrowth2004}{2004}),
who in their seminal paper on the determinants of economic growth,
compile a cross-country data set comprising GDP per capita growth rates
between 1960-1996 for a sample of 88 countries, alongside 67 potential
explanatory variables. The data set has become a workhorse for testing
high-dimensional regression techniques, particularly in the Bayesian
literature
(\protect\hyperlink{ref-eicherDefaultPriorsPredictive2011}{Eicher
\emph{et al.}, 2011};
\protect\hyperlink{ref-hofmarcherFishingEconomicGrowth2011}{Hofmarcher
\emph{et al.}, 2011};
\protect\hyperlink{ref-leyEffectPriorAssumptions2008}{Ley, 2008};
\protect\hyperlink{ref-sala-i-martinDeterminantsLongTermGrowth2004}{Sala-I-Martin
\emph{et al.}, 2004};
\protect\hyperlink{ref-schneiderCatchingGrowthDeterminants2012}{Schneider
\& Wagner, 2012}). The econometric techniques that have been employed
include Bayesian model averaging, as well as various model selection and
shrinkage methods such as the ElasticNet and Lasso estimators. A
description of the variables contained in the data set is provided in
Table \ref{tab:vars}.\footnote{Since the HFR as well as benchmark
  methods require the ranges of the input variables to be similar, the
  67 predictors in the data set are scaled to an interval of \([-1,1]\).
  Dummy variables are normalized to a range of \([-0.5, 0.5]\). This is
  done to dampen the otherwise overstated effect of the dummy variables
  in the hierarchy. The GDP per capita growth variable is not
  transformed to ensure that a comparison to previous research is
  possible. All specifications discussed in this section include an
  intercept term.}

\begin{figure}[H]

{\centering \includegraphics{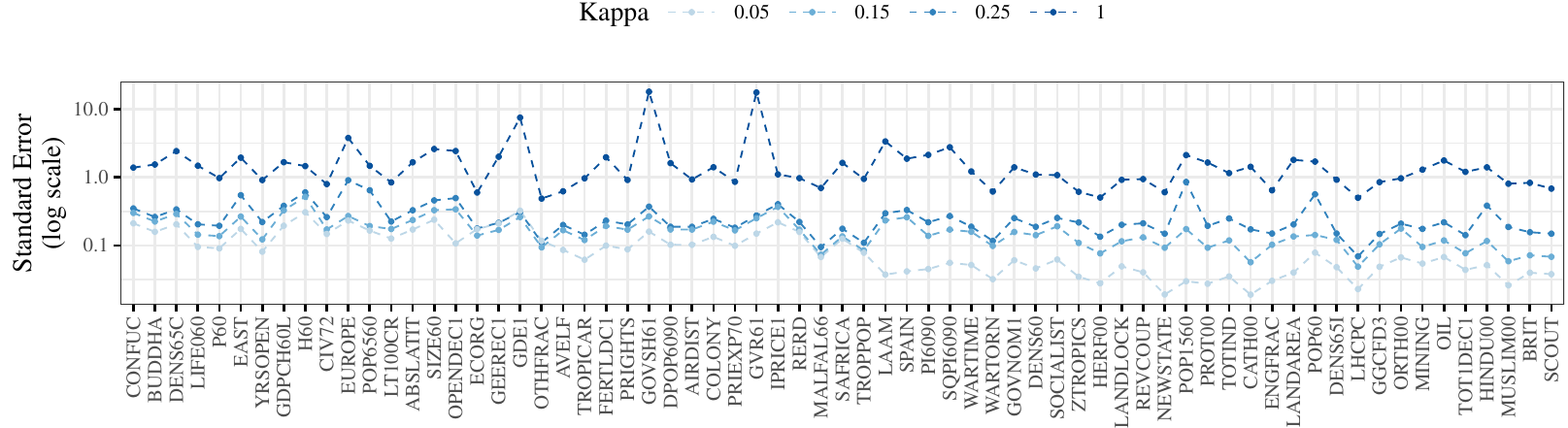} 

}

\caption{Approximate standard errors of the HFR estimates using four different settings for $\kappa$ (log scale). As the bias of the estimates increases with higher $\kappa$, the variance decreases. The standard errors represent weighted averages over the level-specific standard errors as described in Section \ref{method:constant}.\label{fig:kappa_se}}\label{fig:unnamed-chunk-7}
\end{figure}

\begin{landscape}

\begin{figure}[p]

{\centering \includegraphics[angle=0]{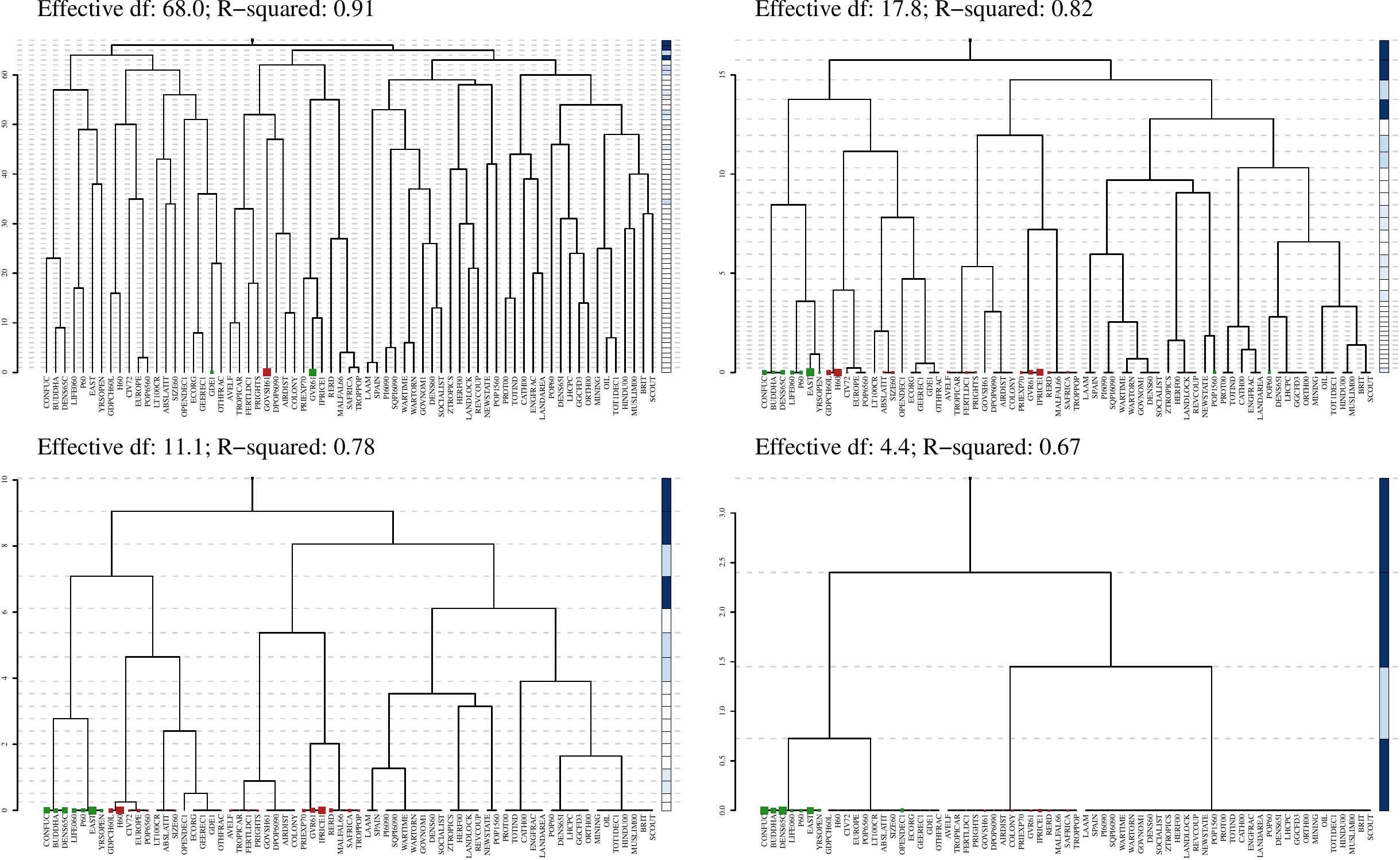} 

}

\caption{Dendrograms for the HFR coefficients obtained from four different settings for $\kappa$. The top-left panel represents the unconstrained case, while the bottom-left panel is the hyperparameter that minimizes a 10-fold cross-validated MSE.\label{fig:graph_unconstrained}}\label{fig:unnamed-chunk-8}
\end{figure}

\end{landscape}

As a starting point, Fig. \ref{fig:graph_unconstrained} depicts
hierarchical graphs for 4 different settings of \(\kappa\) --- the
hyperparameter governing the size of the optimal graph. The
unconstrained regression graph is plotted in the top-left panel, with a
total height of \(67\) (\(\kappa = 1\)) and each level contributing to a
maximum extent (\(\boldsymbol{\theta}=\mathbf{1}\)). The graph is highly
complex, reflecting the dimensionality of the problem, and is difficult
to interpret in a meaningful manner. The regression coefficients
themselves, which are represented by the leaf nodes, are estimated with
substantial variance (see Fig. \ref{fig:kappa_se}), highlighting the
need for a regularized approach. The remaining panels of Fig.
\ref{fig:graph_unconstrained} show different degrees of shrinkage,
leading to successively simpler hierarchies. Each lower value of
\(\kappa\) increases the strength of shrinkage (and hence the parameter
bias), while in turn decreasing the variability of the estimates, as
demonstrated in Fig. \ref{fig:kappa_se}.

\newpage

In contrast to the complex unconstrained structure, Fig.
\ref{fig:graph_optimal} displays the estimated optimal shrinkage tree
for the regression. The height of the tree is 10.1, with
\(\nu_{\text{eff}} = 10.1+1\) determined using a 10-fold
cross-validation procedure. The distance between the levels reflects the
shrinkage weights \(\boldsymbol{\theta}\), and the vertical bar on the
right is shaded based on the contribution of each level to the overall
coefficient of determination of the HFR fit:

\begin{figure}[H]

{\centering \includegraphics{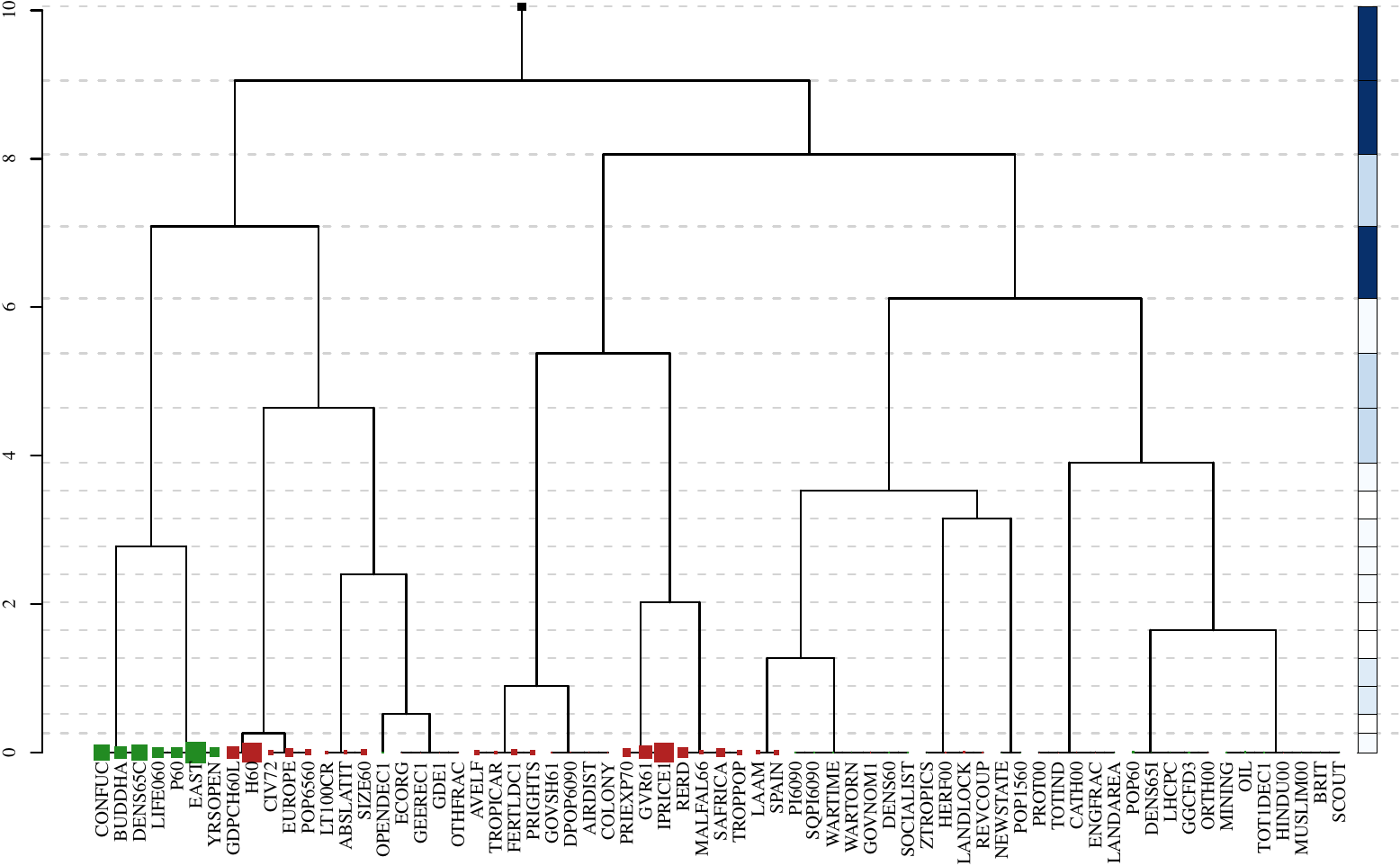} 

}

\caption{Dendrogram for the HFR coefficients for the hyperparameter that minimizes a 10-fold cross-validated MSE.\label{fig:graph_optimal}}\label{fig:unnamed-chunk-9}
\end{figure}

Fig. \ref{fig:graph_optimal} suggests that the primary contribution to
model fit is derived from the upper levels. Examining the level-wise
contributions directly in Fig. \ref{fig:scree} shows that only the first
18 levels contribute to the fitting process and the first four levels
account for over 85\% of the explained variation. The plot is analogous
to scree plots produced for principal components regressions, with the
summation over the level-specific contributions yielding the total
\(R^2\) of the HFR fit:

\begin{figure}[H]

{\centering \includegraphics{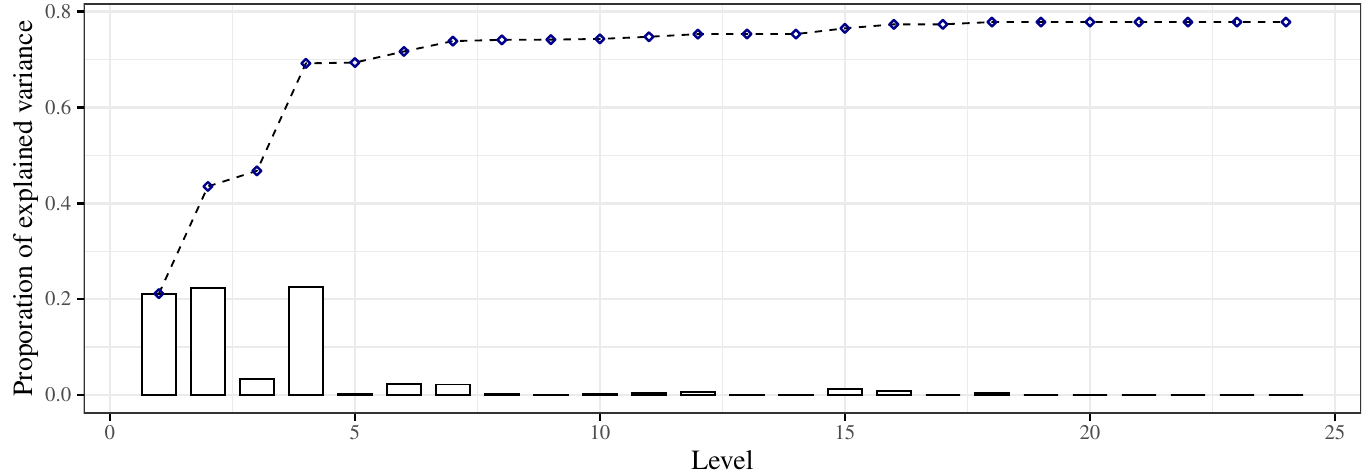} 

}

\caption{Cumulative $R^2$ over the levels in the optimal hierarchy (dashed line), and level-specific contributions to $R^2$ (bars). Only levels $\ell < 25$ are plotted. The remaining levels do not contribute to the model fit.\label{fig:scree}}\label{fig:unnamed-chunk-10}
\end{figure}

As illustrated in the bottom-right panel of Fig.
\ref{fig:graph_unconstrained}, the first four hierarchical levels divide
the sample into four latent signal factors that explain a significant
portion of the response variation. The factors appear to identify
regional or topical sub-clusters, as well as consolidated noise
components. The first cluster contains variables that identify the East
Asian region (e.g.~BUDDHA, CONFUC, EAST). The second cluster appears to
group mostly institutional quality measures and some related variables
(e.g.~H60, CIV72, OPENDEC1, ECORG). The third cluster groups variables
that presumably identify developing economies (e.g.~MALFAL66, SAFRICA,
TROPPOP) and several closely related economic measures (e.g.~RERD,
IPRICE1, PRIEXP70). Finally, the fourth cluster contains a large group
of variables with coefficients close to zero, suggesting that these
measures represent primarily noise components.

The fact that the upper clusters enter with a much higher importance
than their corresponding leaf nodes, may suggest that the common --- as
opposed to the idiosyncratic --- information in the predictor groups
determines growth disparities. For instance, rather than malaria
prevalence entering as a growth determinant in its own right, the
variable (MALFAL66) helps to identify an underlying geographic factor
that drives economic growth.

\begin{figure}[H]

{\centering \includegraphics{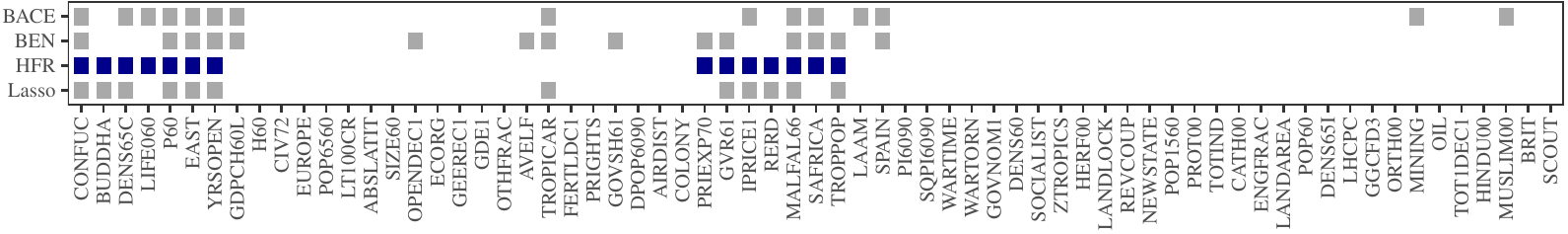} 

}

\caption{Model selection using the approximate $p$-values of the HFR coefficients compared to BACE, BEN and Lasso regressions.\label{fig:model_comp}}\label{fig:unnamed-chunk-11}
\end{figure}

Examining the individual growth drivers more closely, Fig.
\ref{fig:model_comp} displays the most important variables identified by
\protect\hyperlink{ref-sala-i-martinDeterminantsLongTermGrowth2004}{Sala-I-Martin
\emph{et al.}}
(\protect\hyperlink{ref-sala-i-martinDeterminantsLongTermGrowth2004}{2004})
(BACE) and
\protect\hyperlink{ref-hofmarcherFishingEconomicGrowth2011}{Hofmarcher
\emph{et al.}}
(\protect\hyperlink{ref-hofmarcherFishingEconomicGrowth2011}{2011})
(BEN), as well as all HFR coefficients with an indicative
\(p\text{-value}<0.05\).\footnote{The \(p\)-value is calculated using
  average standard errors as described in Section \ref{method:constant}
  with the residual degrees of freedom given by \(N-\nu_{\text{eff}}\).}
As an auxiliary comparison, the model selected using a Lasso estimator
is also displayed.\footnote{As for the HFR, the Lasso penalty is
  determined using a 10-fold cross-validation approach.} The HFR
identifies a total of 14 growth determinants grouped into two blocks:
those associated with cluster one and those associated with cluster
three. The variable set closely resembles related studies (with 12 of 14
overlapping drivers), but reflects the clustering inherent to the HFR.
The model selected using the Lasso is almost identical to the HFR, with
all but one of the growth determinants taken from the two relevant
effect clusters discovered by the HFR.

A key consideration for the validity of the uncovered model and the
quality of the HFR estimates is the method's predictive performance. In
order to assess this systematically, I employ a sampling setup closely
resembling
\protect\hyperlink{ref-hofmarcherFishingEconomicGrowth2011}{Hofmarcher
\emph{et al.}}
(\protect\hyperlink{ref-hofmarcherFishingEconomicGrowth2011}{2011}).
Observations are randomly sampled to form training, validation and
testing sets containing 68/10/10 observations, respectively.\footnote{These
  proportions are roughly equivalent to those used in
  \protect\hyperlink{ref-hofmarcherFishingEconomicGrowth2011}{Hofmarcher
  \emph{et al.}}
  (\protect\hyperlink{ref-hofmarcherFishingEconomicGrowth2011}{2011}),
  but with the addition of a validation sample, which is obtained by
  reducing the size of both the training and testing samples slightly.}
Parameters are estimated using the training sample, hyperparameters are
determined via a grid search minimization of the validation MSE and the
performance is calculated as the test sample MSE. Samples are drawn in
500 iterations with hyperparameters determined independently in each
run.

\begin{figure}[H]

{\centering \includegraphics{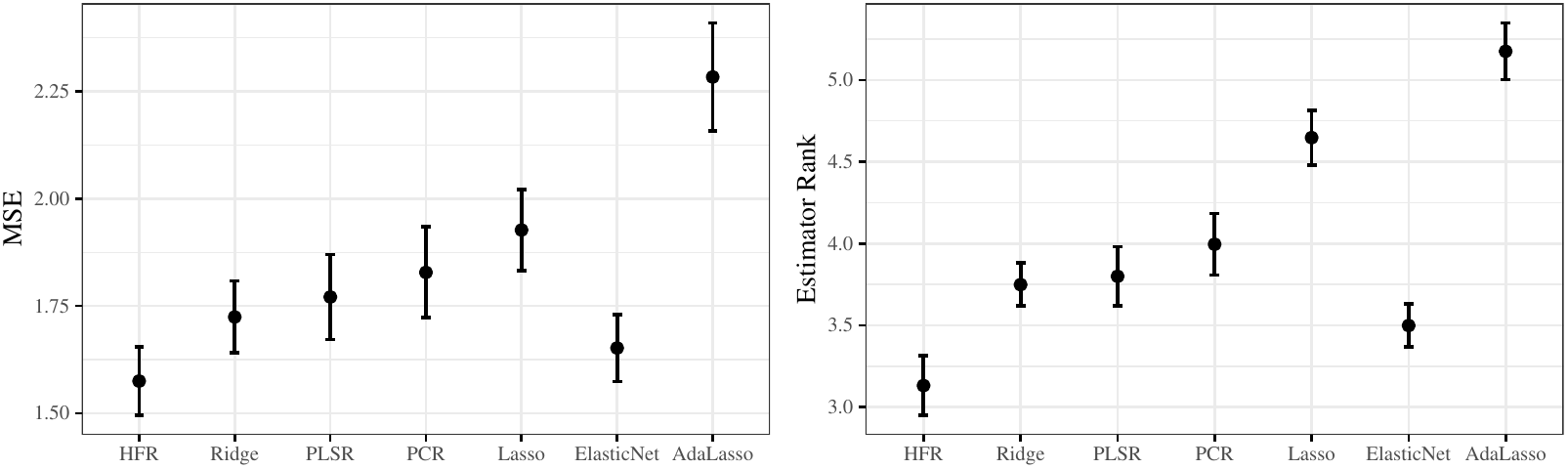} 

}

\caption{Comparison of prediction accuracy of HFR, Ridge, PLSR, PCR, ElasticNet, Lasso, and AdaLasso for GDP per capita growth from 1960-1996. MSE (left panel), rank of estimators (right panel). Statistics plotted as mean and 95\% confidence interval based on 500 random training, validation and testing samples. Prediction errors are multiplied by 1e4. \label{fig:gdp}}\label{fig:unnamed-chunk-12}
\end{figure}

Figure \ref{fig:gdp} plots the MSE and the average rank for the HFR and
a panel of benchmark methods. The benchmark methods include penalized
regressions in the form of the ridge regression, Lasso, Adaptive Lasso
(AdaLasso) and ElasticNet, latent variable regressions in the form of
PCR and PLSR, and finally OLS.\footnote{Ridge, Lasso and ElasticNet are
  implemented using the \texttt{glmnet}-package in the statistical
  computing language \texttt{R}, described in
  \protect\hyperlink{ref-friedmanRegularizationPathsGeneralized2010}{Friedman
  \emph{et al.}}
  (\protect\hyperlink{ref-friedmanRegularizationPathsGeneralized2010}{2010}).
  For a discussion of the AdaLasso, see
  \protect\hyperlink{ref-zouAdaptiveLassoIts2006}{Zou}
  (\protect\hyperlink{ref-zouAdaptiveLassoIts2006}{2006}). PCR and PLSR
  are implemented using the \texttt{pls}-package in the statistical
  computing language \texttt{R}, described in
  \protect\hyperlink{ref-mevikIntroductionPlsPackage2019}{Mevik \&
  Wehrens}
  (\protect\hyperlink{ref-mevikIntroductionPlsPackage2019}{2019}).} In
addition, Table \ref{tab:gdp} displays the distribution of the MSEs
alongside the results of the BACE and the BEN. The estimation of BACE
and BEN is not replicated, but the results are taken directly from the
table presented in
\protect\hyperlink{ref-hofmarcherFishingEconomicGrowth2011}{Hofmarcher
\emph{et al.}}
(\protect\hyperlink{ref-hofmarcherFishingEconomicGrowth2011}{2011}),
page 10.

\begin{table}[ht]
\centering
\begin{tabular}{l|c|cccccc|cc}
  \toprule
 & HFR & Ridge & PLSR & PCR & Lasso & ElasticNet & AdaLasso & BACE & BEN \\ 
  \midrule
Min. & 0.307 & 0.268 & 0.298 & 0.267 & 0.211 & 0.296 & 0.252 & 0.358 & 0.452 \\ 
  1st Qu. & 0.958 & 1.053 & 1.016 & 1.075 & 1.212 & 1.030 & 1.304 & 1.122 & 1.114 \\ 
  Median & 1.329 & 1.543 & 1.486 & 1.502 & 1.708 & 1.489 & 1.934 & 1.614 & 1.494 \\ 
  Mean & 1.575 & 1.724 & 1.771 & 1.828 & 1.927 & 1.652 & 2.283 & 1.705 & 1.587 \\ 
  3rd Qu. & 1.973 & 2.201 & 2.214 & 2.164 & 2.395 & 2.061 & 2.865 & 2.158 & 1.869 \\ 
  Max. & 5.851 & 5.773 & 8.805 & 10.113 & 7.466 & 5.811 & 10.311 & 4.213 & 3.891 \\ 
   \bottomrule
\end{tabular}
\caption{Distribution of prediction error (MSE) for GDP per capital growth based on 500 simulation runs. Prediction errors are multiplied by 1e4.} 
\label{tab:gdp}
\end{table}

The HFR outperforms all benchmark methods, with the ElasticNet
regression achieving the highest mean accuracy among the panel of
non-Bayesian benchmarks in Fig. \ref{fig:gdp}. When compared to the
performance of the BACE and BEN models, the HFR is again found to
achieve lower mean and median prediction errors. The results provide
justification for the approach taken by the HFR, suggesting that the
aggregation of growth determinants into a low-dimensional set of latent
factors is indeed appropriate.

In sum, the HFR offers a dual benefit: (i) it generates robust
out-of-sample predictions, while (ii) the parsimonious hierarchy, in
which the estimates are embedded, produces meta-insights about the
underlying latent signals that explain observed response variation. In
the case of the determinants of economic growth, several regional and
topical sub-clusters may suffice to offer robust explanations of
observed growth disparities. This ability to distinguish between the
types of explanatory variation (shared or idiosyncratic) within a fully
supervised framework is unique to the HFR, and can provide valuable
additional insight with respect to the data generating process. The
following section tests the generality of the observed predictive
accuracy in simulated conditions.

\hypertarget{simulations}{%
\section{\texorpdfstring{Simulations\label{simulations}}{Simulations}}\label{simulations}}

I use four simulations, largely replicated from related work, that cover
different types of regression tasks to compare the performance of the
HFR to similar methods. The benchmark methods include penalized
regressions in the form of the ridge regression, Lasso, Adaptive Lasso
(AdaLasso) and ElasticNet, latent variable regressions in the form of
PCR and PLSR, and finally OLS.\footnote{Ridge, Lasso and ElasticNet are
  implemented using the \texttt{glmnet}-package in the statistical
  computing language \texttt{R}, described in
  \protect\hyperlink{ref-friedmanRegularizationPathsGeneralized2010}{Friedman
  \emph{et al.}}
  (\protect\hyperlink{ref-friedmanRegularizationPathsGeneralized2010}{2010}).
  For a discussion of the AdaLasso, see
  \protect\hyperlink{ref-zouAdaptiveLassoIts2006}{Zou}
  (\protect\hyperlink{ref-zouAdaptiveLassoIts2006}{2006}). PLSR and PCR
  are implemented using the \texttt{pls}-package in the statistical
  computing language \texttt{R}, described in
  \protect\hyperlink{ref-mevikIntroductionPlsPackage2019}{Mevik \&
  Wehrens}
  (\protect\hyperlink{ref-mevikIntroductionPlsPackage2019}{2019}).}

The simulations show that each benchmark method is particularly well
suited to certain regression tasks and poorly to others, as is generally
observed in the related literature
(\protect\hyperlink{ref-tibshiraniRegressionShrinkageSelection1996}{Tibshirani,
1996}). The HFR, by contrast, exhibits a high degree of versatility,
outperforming or matching the benchmark methods in all simulations.
Three of the simulations are based on
\protect\hyperlink{ref-tibshiraniRegressionShrinkageSelection1996}{Tibshirani}
(\protect\hyperlink{ref-tibshiraniRegressionShrinkageSelection1996}{1996})
and \protect\hyperlink{ref-zouRegularizationVariableSelection2005}{Zou
\& Hastie}
(\protect\hyperlink{ref-zouRegularizationVariableSelection2005}{2005})
and have been applied occasionally in similar research
(\protect\hyperlink{ref-bondellSimultaneousRegressionShrinkage2008}{Bondell
\& Reich, 2008}). The final simulation is new.

Data is simulated from the true model \begin{equation}
y = \boldsymbol{x \beta} + \epsilon , \;\;\; \epsilon \sim \mathcal{N}(0,\sigma^2). \nonumber
\end{equation} Observations are divided into training, validation and
testing samples, where the training sample is used to estimate the
models, the validation sample is used to determine optimal
hyperparameters, and the testing sample is used for performance
evaluation. Model performance is assessed by calculating the mean
squared error (MSE) over the testing sample. Sample sizes are denoted by
\(\cdot/\cdot/\cdot\), where the dots represent training, validation and
testing samples, respectively. In each case, the results of 500
simulation runs are plotted.

Hyperparameters include \(\kappa\) for the HFR, the size of the penalty
(\(\lambda\)) for the penalized estimators (ridge, Lasso, AdaLasso,
ElasticNet), the mixing parameter (\(\alpha\)) for the ElasticNet, and
the number of latent components for the PCR and PLSR. Optimal
hyperparameter values are determined using an extensive grid search with
selection based on the minimum validation MSE. Hyperparameter tuning is
performed individually for each method in each simulation run.

\hypertarget{simulation-setup}{%
\subsection{\texorpdfstring{Simulation setup
\label{simulations:setup}}{Simulation setup }}\label{simulation-setup}}

\textbf{Simulation (a)} is taken from
\protect\hyperlink{ref-tibshiraniRegressionShrinkageSelection1996}{Tibshirani}
(\protect\hyperlink{ref-tibshiraniRegressionShrinkageSelection1996}{1996}),
where it was originally used to demonstrate the performance of the ridge
regression. True parameter values are set to
\(\beta_j = 0.85, \; \forall \; j = 1, ..., K\), with \(K = 8\). The
sample size is 20/20/200, \(\sigma^2 = 3\) and the pairwise correlation
between \(x_i\) and \(x_j\) is \(0.5^{|i-j|}\).

\textbf{Simulation (b)} is again based on
\protect\hyperlink{ref-tibshiraniRegressionShrinkageSelection1996}{Tibshirani}
(\protect\hyperlink{ref-tibshiraniRegressionShrinkageSelection1996}{1996})
and is a sparse regression used to illustrate the Lasso's ability of
eliminating noise features. There are 40 predictors with parameters set
to
\[\boldsymbol{\beta} = (\underset{10}{\underbrace{0,...,0}},\underset{10}{\underbrace{2,...,2}},\underset{10}{\underbrace{0,...,0}},\underset{10}{\underbrace{2,...,2}}).\]
As before the pairwise correlation between \(x_i\) and \(x_j\) is
\(0.5^{|i-j|}\), and \(\sigma^2 = 15\). The sample size is set to
\(100/100/400\). Since the DGP is sparse, the task is likely to be
solved well with a Lasso, AdaLasso or ElasticNet. Variations on this
simulation are used in Section \ref{simulations:violating} to explore
scenarios for which the HFR is less suitable.

\textbf{Simulation (c)} is taken from
\protect\hyperlink{ref-zouRegularizationVariableSelection2005}{Zou \&
Hastie}
(\protect\hyperlink{ref-zouRegularizationVariableSelection2005}{2005}),
who study the effect of grouped predictors. The simulation contains a
mixture of grouped predictors and noise predictors and is therefore a
grouped feature selection task. The sample consists of 50/50/400
observations and 40 predictors with
\[\boldsymbol{\beta} = (\underset{15}{\underbrace{3,...,3}},\underset{25}{\underbrace{0,...,0}}),\]
\(\sigma^2 = 15\), and \(\mathbf{x}\) generated as follows (with
\(\epsilon^x_i \sim \mathcal{N}(0, 0.01)\)): \begin{align}
  &x_i = \xi_1 + \epsilon^x_i, \;\;\; \xi_1 \sim \mathcal{N}(0,1), \;\;\; i = 1,...,5 \nonumber \\
  &x_i = \xi_2 + \epsilon^x_i, \;\;\; \xi_2 \sim \mathcal{N}(0,1), \;\;\; i = 6,...,10 \nonumber \\
  &x_i = \xi_3 + \epsilon^x_i, \;\;\; \xi_3 \sim \mathcal{N}(0,1), \;\;\; i = 11,...,15 \nonumber \\
  &x_i \sim \mathcal{N}(0,1), \;\;\; i = 16,...,40. \nonumber 
  \end{align} The simulation is designed to illustrate the ability of
the ElasticNet to deal with grouped variables and variable selection
simultaneously, and should therefore see the ElasticNet performing well.

\textbf{Simulation (d)} is designed to test predictive performance in
the presence of latent factors. The sample consists of 20/20/200
observations. Simulation (d) draws from a true model
\(y = \boldsymbol{f}\boldsymbol{\beta} + \epsilon\) where
\(\boldsymbol{f} = \begin{bmatrix} f_1 & \cdots & f_4 \end{bmatrix}\),
\(\boldsymbol{\beta} = \begin{pmatrix}1.0,1.5,2.0,1.5\end{pmatrix}\),
\(\sigma^2 = 3\) and the pairwise correlation between \(f_i\) and
\(f_j\) is \(0.5^{|i-j|}\). Unlike the previous cases, I assume
\(\mathbf{x}\) contains noisy measures of the unobserved latent factors
\(\boldsymbol{f}\), such that (with
\(\epsilon^x_i \sim \mathcal{N}(0, 1)\)): \begin{align}
  &x_i = f_1 + f_2 + \epsilon^x_i, \;\;\; i = 1,2 \nonumber \\
  &x_i = f_2 + f_3 + \epsilon^x_i, \;\;\; i = 3,4 \nonumber \\
  &x_i = f_3 + f_4 + \epsilon^x_i, \;\;\; i = 5,6 \nonumber \\
  &x_i = f_4 + f_1 + \epsilon^x_i, \;\;\; i = 7,8 \nonumber
  \end{align} The PCR is expected to outperform other regularized
regressions in this example.

\hypertarget{simulation-results}{%
\subsection{Simulation results}\label{simulation-results}}

\begin{figure}[H]

{\centering \includegraphics{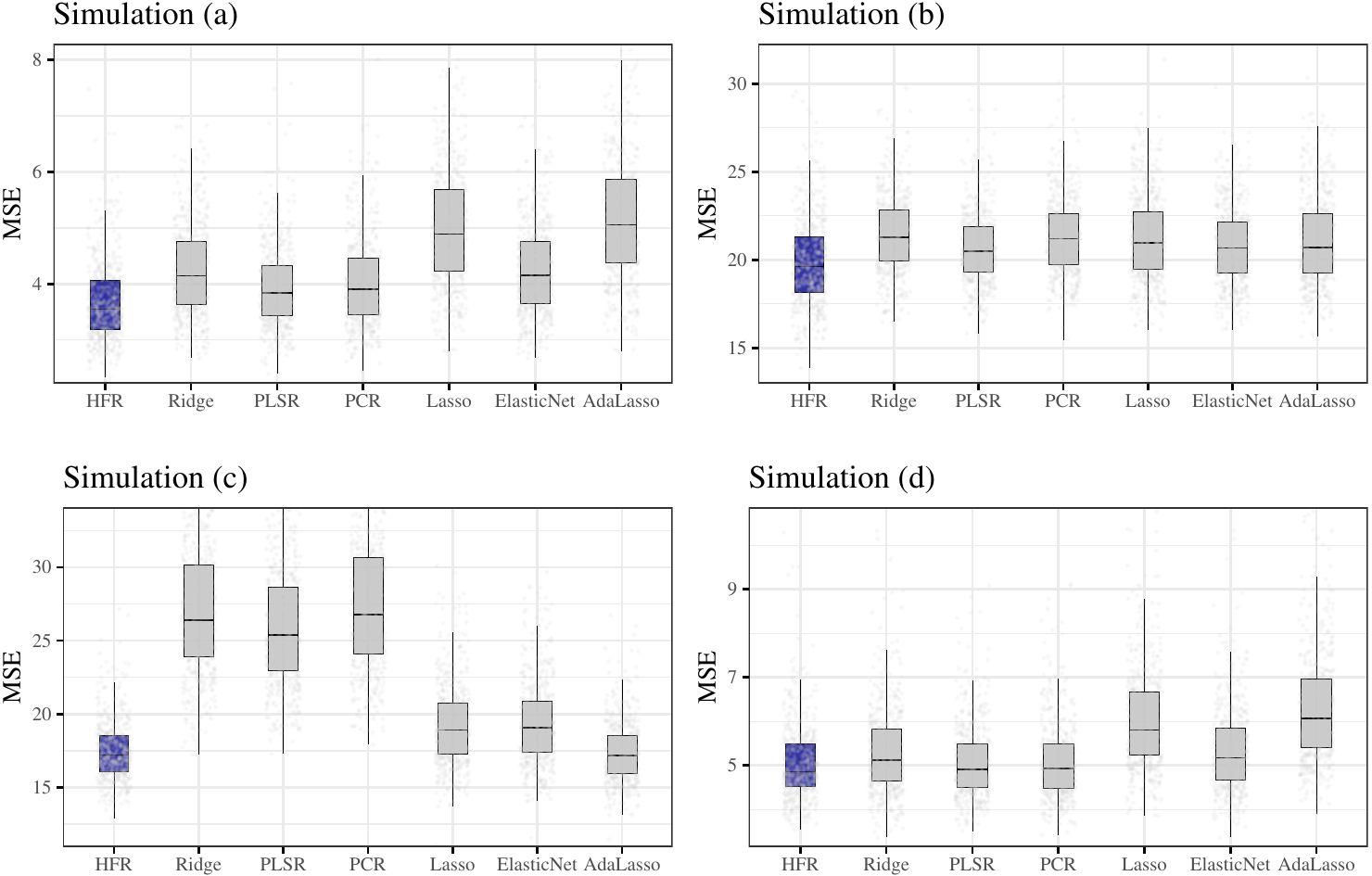} 

}

\caption{Comparison of prediction accuracy of hierarchical feature regression (HFR), Ridge, PLSR, PCR, ElasticNet, Lasso and AdaLasso for simulations (a)-(d). \label{fig:sim_pred}}\label{fig:unnamed-chunk-14}
\end{figure}

Figure \ref{fig:sim_pred} plots the model accuracy for Simulations (a)
to (d). The HFR outperforms or closely matches the benchmarks in all
simulations. Good performance in the cases when no predictor groups
exist in the true DGP (Simulations (a) \& (b)), or when an overlapping
grouping structure exists (Simulation (d)) illustrate the versatility of
the HFR in estimating robust parameters. The feature selection tasks
(Simulations (b) \& (c)) demonstrate how the ability to group noise
features can lead to good performance even when compared to methods that
explicitly perform variable selection, such as the Lasso, AdaLasso and
ElasticNet regressions.

\begin{figure}[H]

{\centering \includegraphics{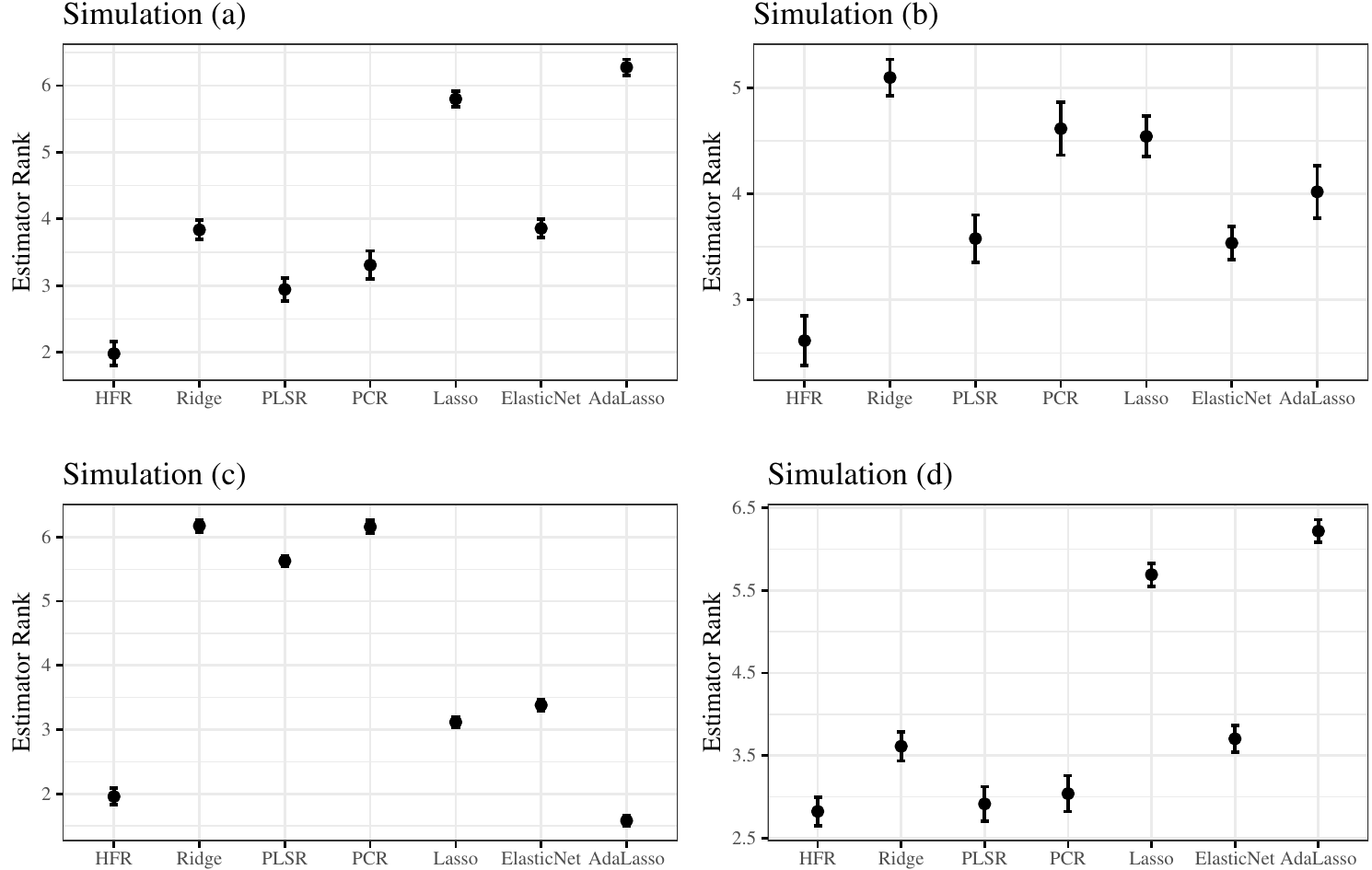} 

}

\caption{Comparison of average predictor rank by MSE of HFR, Ridge, PLSR, PCR, ElasticNet, Lasso and AdaLasso for simulations (a)-(d). Mean and 95\% confidence interval from 500 simulation runs are plotted. \label{fig:sim_pred_rank}}\label{fig:unnamed-chunk-15}
\end{figure}

The average comparative rank of the different estimators is plotted in
Figure \ref{fig:sim_pred_rank} and suggests a highly favorable relative
performance of the HFR, with the lowest or second lowest mean rank
achieved in each instance. The figure is useful in uncovering relative
performance attributes not easily discerned in Figure
\ref{fig:sim_pred}, such as the superior accuracy of the ElasticNet in
Simulation (b) in relation to other feature selection algorithms like
Lasso or AdaLasso, and provides additional evidence of the accuracy and
versatility of the HFR.

Table \ref{tab:sim_pred} summarizes the results of the simulations
including bootstrap standard errors for the median MSE performance
metrics:

\begin{table}[ht]
\centering
\begin{tabular}{lcccc}
  \toprule
 & Sim. (a) & Sim. (b) & Sim. (c) & Sim. (d) \\ 
  \midrule
HFR & \textbf{3.546} (0.025) & \textbf{19.616} (0.144) & \textbf{17.205} (0.119) & \textbf{4.862} (0.043) \\ 
  Ridge & 4.146 (0.062) & 21.28 (0.097) & 26.408 (0.201) & 5.118 (0.05) \\ 
  PLSR & \textbf{3.837} (0.039) & \textbf{20.491} (0.09) & 25.391 (0.247) & \textbf{4.907} (0.052) \\ 
  PCR & 3.909 (0.054) & 21.191 (0.098) & 26.781 (0.233) & 4.93 (0.03) \\ 
  Lasso & 4.89 (0.056) & 20.956 (0.091) & 18.919 (0.091) & 5.805 (0.054) \\ 
  ElasticNet & 4.151 (0.053) & 20.669 (0.116) & 19.085 (0.129) & 5.171 (0.048) \\ 
  AdaLasso & 5.059 (0.057) & 20.699 (0.155) & \textbf{17.162} (0.105) & 6.067 (0.074) \\ 
  OLS & 5.4 (0.11) & 25.15 (0.138) & 81.851 (1.67) & 7.318 (0.132) \\ 
   \bottomrule
\end{tabular}
\caption{Prediction accuracy (median MSE) for simulations (a)-(d) based on 500 simulation runs. Standard errors in parantheses. Standard errors are calculated using 500 bootstrap resamplings of the estimated MSE. In each case the two best methods are highlighted.} 
\label{tab:sim_pred}
\end{table}

\hypertarget{trace-plots}{%
\subsection{\texorpdfstring{Trace
plots\label{method:trace}}{Trace plots}}\label{trace-plots}}

\begin{figure}[H]

{\centering \includegraphics{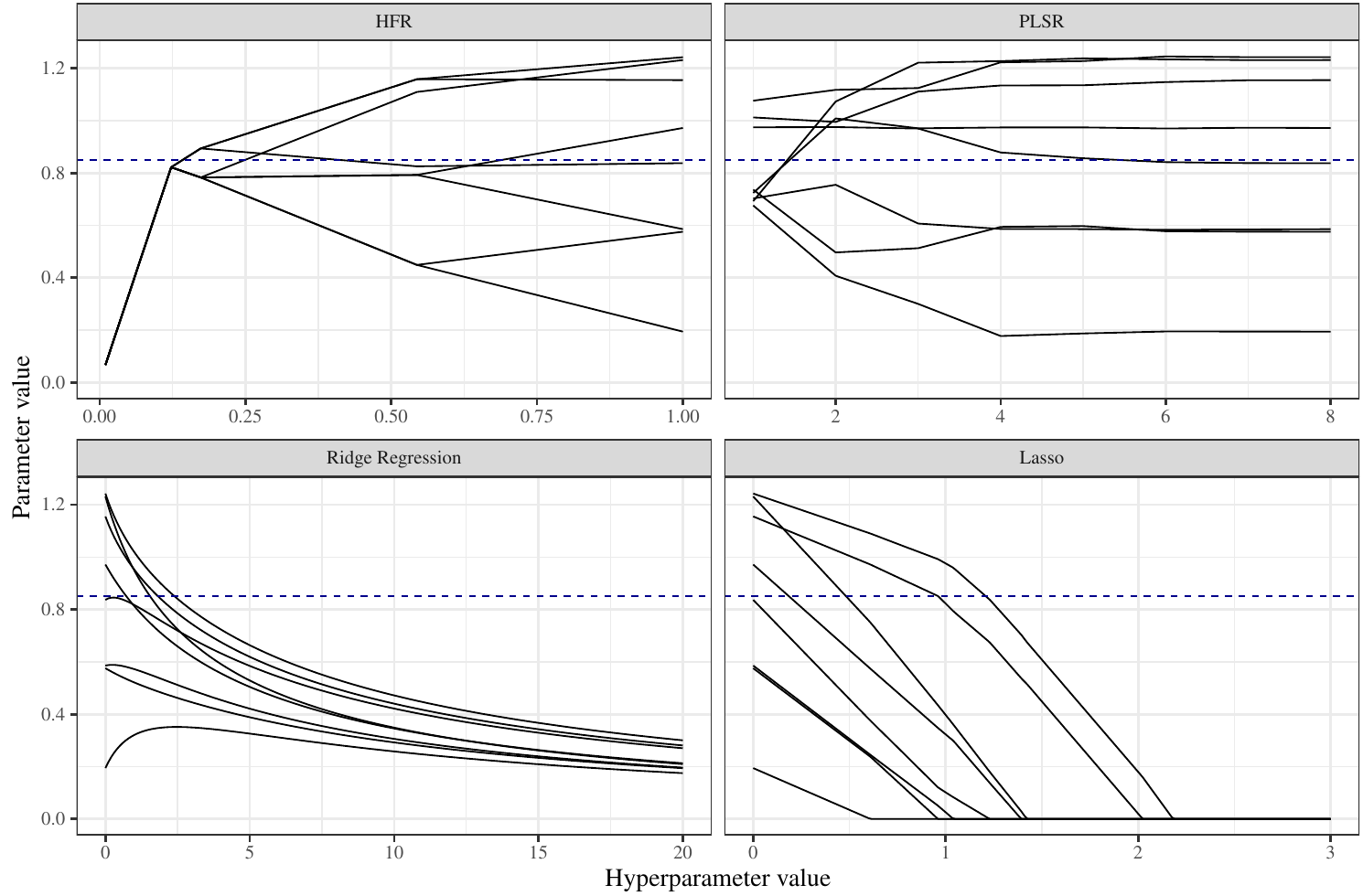} 

}

\caption{Trace plots for HFR, PLSR, ridge and Lasso for $K=8$ predictors. The hyperparameter values are $\kappa$ for the HFR, the number of components for the PLSR, and $\lambda$ for ridge and Lasso. True parameter values are represented by the dashed line. \label{fig:trplots}}\label{fig:unnamed-chunk-17}
\end{figure}

In order to explore shrinkage behavior in the HFR, Fig.
\ref{fig:trplots} (top-left panel) draws trace plots of
\(\boldsymbol{\hat{\beta}}_{\text{hfr}}\), using the setup in Simulation
(a), with \(K = 8\) predictors. The plot illustrates, how parameter
estimates are drawn towards group targets as \(\kappa\) decreases. The
estimates are eventually shrunken towards zero for very small values of
\(\kappa\).

By way of comparison, Fig. \ref{fig:trplots} (bottom and upper right
panels) draws trace plots for the PLSR, ridge regression and Lasso
estimators using the same regression problem as above. The plots
highlight the key difference between the HFR and traditional regularized
regressions. The HFR can reduce noise in a highly efficient manner, with
only limited attenuation bias in the estimates (see dashed line in Fig.
\ref{fig:trplots}). For the given regression problem, ridge and Lasso,
which penalize the parameter norm and reduce all coefficient estimates
towards zero, can only eliminate a meaningful portion of the noise at
levels of \(\lambda\) that result in parameters estimated with
substantial attenuation bias.

\hypertarget{boundary-cases}{%
\subsection{\texorpdfstring{Boundary cases
\label{simulations:violating}}{Boundary cases }}\label{boundary-cases}}

While the above results demonstrate the high degree of accuracy and
versatility of the HFR, it can potentially yield less convincing
outcomes in at least two scenarios, as illustrated in the following
variations on Simulation (b):

\textbf{Simulation (e)} repeats Simulation (b), but with true parameters
equal to
\[\boldsymbol{\beta} = (\underset{10}{\underbrace{0,...,0}},\underset{10}{\underbrace{1,...,3}},\underset{10}{\underbrace{0,...,0}},\underset{10}{\underbrace{-1,...,-3}}).\]
Here the second and fourth parameter blocks represent evenly spaced
sequences on the interval \([1,3]\) and \([-3,-1]\), respectively. This
requires the HFR to construct a shallower hierarchy, since the effect of
the predictors on \(y\) is more heterogeneous and more idiosyncratic
information must be included. A shallower hierarchy limits the feasible
extent of regularization, and results in a higher effective degrees of
freedom.

\textbf{Simulation (f)} is identical to Simulation (b), but with a
pairwise correlation between \(x_i\) and \(x_j\) of 0.5. The high degree
of correlation between noise and signal predictors results in an
extremely noisy \(\boldsymbol{\mathcal{D}}_y\) matrix that cannot be
clustered in any meaningful manner. Thus, the HFR can only poorly
distinguish between signal and noise predictors and hierarchy
construction becomes essentially random.

\begin{figure}[H]

{\centering \includegraphics{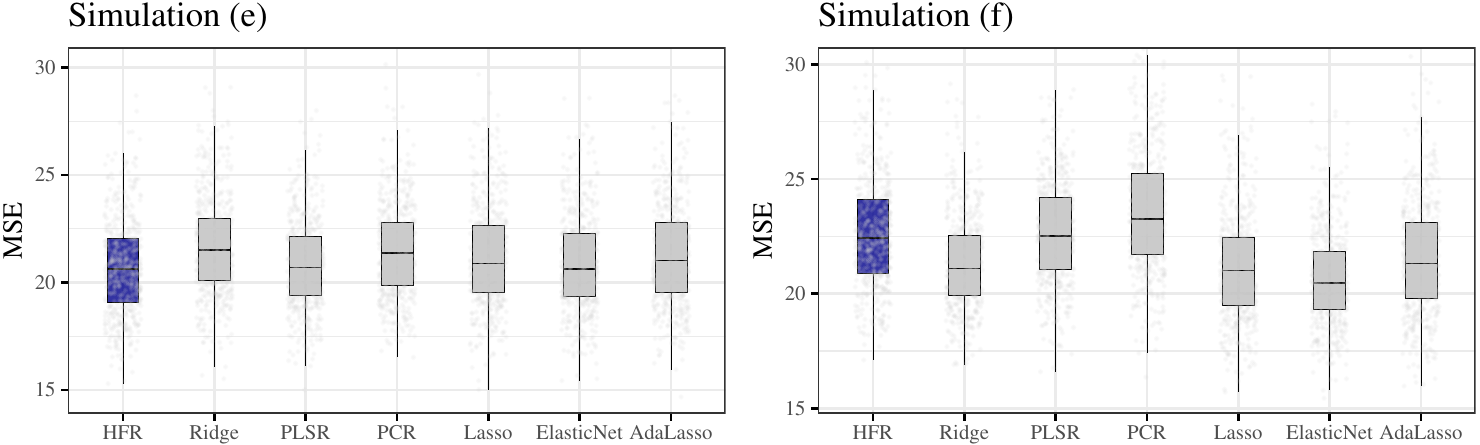} 

}

\caption{Comparison of prediction accuracy of hierarchical feature regression (HFR), Ridge, PLSR, PCR, ElasticNet, Lasso, and AdaLasso for simulations (e)-(f) \label{fig:sim_pred_add}}\label{fig:unnamed-chunk-18}
\end{figure}

The results are plotted in Fig. \ref{fig:sim_pred_add} and Table
\ref{tab:sim_pred_add}. In Simulation (e), the variability of all
methods increases, however, HFR again outperforms the benchmarks,
suggesting that the method can achieve good out-of-sample results, even
when the scope for shrinkage is reduced. In Simulation (f), the HFR
exhibits an average performance, roughly on par with the PCR and PLSR,
but worse than penalized regressions, suggesting that significant value
is added by a meaningful clustering of predictors into hierarchical
groups.

\begin{table}[ht]
\centering
\begin{tabular}{lcc}
  \toprule
 & Sim. (e) & Sim. (f) \\ 
  \midrule
HFR & \textbf{20.63} (0.106) & 22.434 (0.125) \\ 
  Ridge & 21.509 (0.137) & 21.113 (0.128) \\ 
  PLSR & 20.692 (0.113) & 22.529 (0.1) \\ 
  PCR & 21.365 (0.13) & 23.254 (0.155) \\ 
  Lasso & 20.878 (0.126) & \textbf{21.013} (0.128) \\ 
  ElasticNet & \textbf{20.62} (0.153) & \textbf{20.467} (0.082) \\ 
  AdaLasso & 21.034 (0.185) & 21.316 (0.135) \\ 
  OLS & 25.15 (0.147) & 25.15 (0.138) \\ 
   \bottomrule
\end{tabular}
\caption{Prediction accuracy (median MSE) for simulations (e)-(f) based on 500 simulation runs. Standard errors in parantheses. Standard errors are calculated using 500 bootstrap resamplings of the estimated MSE. In each case the two best methods are highlighted.} 
\label{tab:sim_pred_add}
\end{table}

In sum, the simulations make a compelling case for the use of the HFR
estimator. The versatility of the method across a spectrum of different
types of regression tasks is a key strength when compared to the
benchmarks, which are typically tailored to serve specialized purposes.

\hypertarget{concluding-remarks}{%
\section{\texorpdfstring{Concluding remarks
\label{conclusion}}{Concluding remarks }}\label{concluding-remarks}}

Prediction tasks with high-dimensional multicollinear predictor sets are
challenging for least squares based fitting procedures, and a large,
productive literature exists advancing various regularized approaches to
addressing the issue. The HFR is a novel contribution to this body of
knowledge, presenting a method of shrinking coefficients towards group
targets along the branches of an optimal predictor graph. Given a
hyperparameter, which is conveniently interpreted as the effective model
size and bounded between 0 and 1, the HFR is able to estimate both a
supervised graph, as well as the optimal regularized coefficients
associated with that graph.

The characteristics of the HFR make it particularly well-suited to
regression applications with an underlying hierarchical or grouped data
generating process, such as high-dimensional factor modeling in
econometric analysis (e.g.~nowcasting with dynamic factor models) or in
finance (e.g.~multi-factor asset pricing). Applications similar to the
gene selection problem discussed in
\protect\hyperlink{ref-zouRegularizationVariableSelection2005}{Zou \&
Hastie}
(\protect\hyperlink{ref-zouRegularizationVariableSelection2005}{2005})
may also prove particularly suitable to the HFR. The ability to plot the
estimated hierarchy and explore the effect of individual clusters or
levels in the regression provides a wealth of auxiliary insights into
the underlying effect structure.

Both the empirical case study and the simulations presented in this
paper suggest that the HFR provides an interesting complement to widely
used regularized regression algorithms such as the Lasso or PLS
regressions. The HFR achieves lower out-of-sample prediction errors than
a panel of benchmark methods across a spectrum of different regression
tasks, making it interesting both in terms of its performance as well as
its versatility. The method can be thought of as a structured hybrid
between a penalized regression and a supervised latent factor
regression, with some benefits of both classes of algorithms, with
potentially good performance across a wider range of data generating
processes.

\newpage

\hypertarget{references}{%
\section*{References}\label{references}}
\addcontentsline{toc}{section}{References}

\hypertarget{refs}{}
\begin{CSLReferences}
\leavevmode\vadjust pre{\hypertarget{ref-bachStructuredSparsityConvex2012}{}}%
Bach, F., Jenatton, R., Mairal, J. \& Obozinski, G. 2012. Structured
{Sparsity} through {Convex Optimization}. \emph{Statistical Science}.
27(4):450--468.

\leavevmode\vadjust pre{\hypertarget{ref-bairPredictionSupervisedPrincipal2006}{}}%
Bair, E., Hastie, T., Paul, D. \& Tibshirani, R. 2006. Prediction by
{Supervised Principal Components}. \emph{Journal of the American
Statistical Association}. 101(473):119--137.

\leavevmode\vadjust pre{\hypertarget{ref-bondellSimultaneousRegressionShrinkage2008}{}}%
Bondell, H.D. \& Reich, B.J. 2008. Simultaneous {Regression Shrinkage},
{Variable Selection} and {Clustering} of {Predictors} with {OSCAR}.
\emph{Biometrics}. 64(1):115--123.

\leavevmode\vadjust pre{\hypertarget{ref-burnhamMultimodel2004}{}}%
Burnham, K.P. \& Anderson, D.R. 2004. Multimodel {Inference} ---
{Understanding} {AIC} and {BIC} in {Model} {Selection}.
\emph{Sociological Methods \& Research}. 33(2):261--304.

\leavevmode\vadjust pre{\hypertarget{ref-dimatteoInterestRatesCluster2004}{}}%
Di Matteo, T., Aste, T. \& Mantegna, R.N. 2004. An {Interest Rates
Cluster Analysis}. \emph{Physica A: Statistical Mechanics and its
Applications}. 339(1-2):181--188.

\leavevmode\vadjust pre{\hypertarget{ref-dieboldMeasuringDynamicsGlobal2015}{}}%
Diebold, F.X. \& Yilmaz, K. 2015. Measuring the {Dynamics} of {Global
Business Cycle Connectedness}. in \emph{Unobserved {Components} and
{Time Series Econometrics}} Illustrated ed. S.J. Koopman \& N. Shephard
(eds.). {Oxford University Press} S.J. Koopman \& N. Shephard (eds.).
45--70.

\leavevmode\vadjust pre{\hypertarget{ref-efronLeastAngleRegression2004}{}}%
Efron, B., Hastie, T., Johnstone, I. \& Tibshirani, R. 2004. Least
{Angle Regression}. \emph{Annals of Statistics}. 32(2):407--499.

\leavevmode\vadjust pre{\hypertarget{ref-eicherDefaultPriorsPredictive2011}{}}%
Eicher, T.S., Papageorgiou, C. \& Raftery, A.E. 2011. Default {Priors}
and {Predictive Performance} in {Bayesian Model Averaging}, with
{Application} to {Growth Determinants}. \emph{Journal of Applied
Econometrics}. 26(1):30--55.

\leavevmode\vadjust pre{\hypertarget{ref-epshteinFeatureHierarchiesObject2005}{}}%
Epshtein, B. \& Uliman, S. 2005. Feature {Hierarchies} for {Object
Classification}. in \emph{Tenth {IEEE International Conference} on
{Computer Vision} ({ICCV}'05) {Volume} 1} {Beijing, China}: {IEEE}.
220--227 Vol. 1.

\leavevmode\vadjust pre{\hypertarget{ref-everitt_cluster_2011}{}}%
Everitt, B., Landau, S., Stahl, D. \& Leese, M. 2011. \emph{Cluster
analysis}. 5th ed ed. (Wiley series in probability and statistics).
Chichester, West Sussex, U.K: Wiley.

\leavevmode\vadjust pre{\hypertarget{ref-friedmanElementsStatisticalLearning2001}{}}%
Friedman, J., Hastie, T. \& Tibshirani, R. 2001. \emph{The {Elements} of
{Statistical Learning}}. First ed. Vol. 1. {Springer series in
statistics Springer, Berlin}.

\leavevmode\vadjust pre{\hypertarget{ref-friedmanRegularizationPathsGeneralized2010}{}}%
Friedman, J., Hastie, T. \& Tibshirani, R. 2010. Regularization {Paths}
for {Generalized Linear Models} via {Coordinate Descent}. \emph{Journal
of Statistical Software}. 33(1).

\leavevmode\vadjust pre{\hypertarget{ref-girshickRichFeatureHierarchies2014}{}}%
Girshick, R., Donahue, J., Darrell, T. \& Malik, J. 2014. Rich {Feature
Hierarchies} for {Accurate Object Detection} and {Semantic
Segmentation}. in \emph{2014 {IEEE Conference} on {Computer Vision} and
{Pattern Recognition}} {Columbus, OH, USA}: {IEEE}. 580--587.

\leavevmode\vadjust pre{\hypertarget{ref-hansenLeastSquaresModel2007}{}}%
Hansen, B.E. 2007. Least {Squares Model Averaging}. \emph{Econometrica}.
75(4):1175--1189.

\leavevmode\vadjust pre{\hypertarget{ref-hansenEconometrics2019}{}}%
Hansen, B.E. 2019. \emph{Econometrics}. Draft ed. {University of
Wisconsin}.

\leavevmode\vadjust pre{\hypertarget{ref-hoerlApplicationRidgeAnalysis1962}{}}%
Hoerl, A.E. 1962. Application of {Ridge Analysis} to {Regression
Problems}. \emph{Chemical Engineering Progress}. 58(3):54--59.

\leavevmode\vadjust pre{\hypertarget{ref-hoerlRidgeRegressionBiased1970}{}}%
Hoerl, A.E. \& Kennard, R.W. 1970. Ridge {Regression}: {Biased
Estimation} for {Nonorthogonal Problems}. \emph{Technometrics}.
12(1):55--67.

\leavevmode\vadjust pre{\hypertarget{ref-hofmarcherFishingEconomicGrowth2011}{}}%
Hofmarcher, P., Cuaresma, J.C., Grun, B. \& Hornik, K. 2011.
\emph{Fishing economic growth determinants using bayesian elastic nets}.
(Research Report Series 113). {Wirtschaftsuniversität WIen}: {Institute
for Statistics and Mathematics}.

\leavevmode\vadjust pre{\hypertarget{ref-jamesEstimationQuadraticLoss1961}{}}%
James, W. \& Stein, C. 1961. Estimation with {Quadratic Loss}.
\emph{Proceedings of the Fourth Berkeley Symposium on Mathematical
Statistics and Probability}. 1:361--380.

\leavevmode\vadjust pre{\hypertarget{ref-jolliffePrincipalComponentAnalysis2002}{}}%
Jolliffe, I.T. 2002. \emph{Principal {Component Analysis}}. Second ed.
(Springer series in statistics). {New York}: {Springer}.

\leavevmode\vadjust pre{\hypertarget{ref-kaufmanFindingGroupsData2005}{}}%
Kaufman, L. \& Rousseeuw, P.J. 2005. \emph{Finding {Groups} in {Data}:
{An Introduction} to {Cluster Analysis}}. First ed. (Wiley series in
probability and mathematical statistics). {Hoboken, N.J}: {Wiley}.

\leavevmode\vadjust pre{\hypertarget{ref-koseInternationalBusinessCycles2003}{}}%
Kose, M.A., Otrok, C. \& Whiteman, C.H. 2003. International {Business
Cycles}: {World}, {Region}, and {Country}-{Specific Factors}. \emph{The
American Economic Review}. 93(4).

\leavevmode\vadjust pre{\hypertarget{ref-leonClusteringAlgorithmsRiskAdjusted2017}{}}%
León, D., Aragón, A., Sandoval, J., Hernández, G., Arévalo, A. \& Niño,
J. 2017. Clustering {Algorithms} for {Risk}-{Adjusted Portfolio
Construction}. \emph{Procedia Computer Science}. 108(C):1334--1343.

\leavevmode\vadjust pre{\hypertarget{ref-leyEffectPriorAssumptions2008}{}}%
Ley, E. 2008. \emph{On the effect of prior assumptions in bayesian model
averaging with applications to growth regression}. (MPRA Papers 6773).

\leavevmode\vadjust pre{\hypertarget{ref-lopezdepradoBuildingDiversifiedPortfolios2016}{}}%
Lopez de Prado, M. 2016. Building {Diversified Portfolios} that
{Outperform Out}-of-{Sample}. \emph{SSRN Electronic Journal}.

\leavevmode\vadjust pre{\hypertarget{ref-maechlerClusterClusterAnalysis2019}{}}%
Maechler, M., Rousseeuw, P., Struyf, A. \& Hornik, K. 2019.
\emph{Cluster: {Cluster Analysis Basics} and {Extensions}}. (R Package
Version 2.1.0).

\leavevmode\vadjust pre{\hypertarget{ref-maimonDataMiningKnowledge2010}{}}%
Maimon, O. \& Rokach, L. 2010. \emph{Data {Mining} and {Knowledge
Discovery Handbook}}. Second ed. {Boston, MA}: {Springer US}.

\leavevmode\vadjust pre{\hypertarget{ref-mallowsCommentsCP1973}{}}%
Mallows, C.L. 1973. Some {Comments} on {CP}. \emph{Technometrics}.
15(4):661--675.

\leavevmode\vadjust pre{\hypertarget{ref-mantegnaHierarchicalStructureFinancial1999a}{}}%
Mantegna, R.N. 1999. Hierarchical {Structure} in {Financial Markets}.
\emph{The European Physical Journal B-Condensed Matter and Complex
Systems}. 11(1):193--197.

\leavevmode\vadjust pre{\hypertarget{ref-martensReliableRelevantModelling2001}{}}%
Martens, H. 2001. Reliable and {Relevant Modelling} of {Real World
Data}: {A Personal Account} of the {Development} of {PLS Regression}.
\emph{Chemometrics and Intelligent Laboratory Systems}. 58(2):85--95.

\leavevmode\vadjust pre{\hypertarget{ref-mevikIntroductionPlsPackage2019}{}}%
Mevik, B.-H. \& Wehrens, R. 2019. Introduction to the pls {Package}.
\emph{R package manuals}.

\leavevmode\vadjust pre{\hypertarget{ref-mishraDeepMachineLearning2017}{}}%
Mishra, C. \& Gupta, D.L. 2017. Deep {Machine Learning} and {Neural
Networks}: {An Overview}. \emph{IAES International Journal of Artificial
Intelligence (IJ-AI)}. 6(2).

\leavevmode\vadjust pre{\hypertarget{ref-pfitzingerHFR2022}{}}%
Pfitzinger, J. 2022. \emph{HFR: Estimate hierarchical feature regression
models}. (R Package Version 0.5.0).

\leavevmode\vadjust pre{\hypertarget{ref-pfitzingerConstrainedHierarchicalRisk2019}{}}%
Pfitzinger, J. \& Katzke, N. 2019. \emph{A constrained hierarchical risk
parity algorithm with cluster-based capital allocation}. (Working Paper
14/2019). Stellenbosch University, Department of Economics.

\leavevmode\vadjust pre{\hypertarget{ref-rcoreteamLanguageEnvironmentStatistical2018}{}}%
R Core Team. 2018. \emph{R: {A Language} and {Environment} for
{Statistical Computing}}. {Vienna, Austria}: {R Foundation for
Statistical Computing}.

\leavevmode\vadjust pre{\hypertarget{ref-raffinotHierarchicalClusteringBased2016}{}}%
Raffinot, T. 2016. Hierarchical {Clustering} based {Asset Allocation}.
\emph{SSRN Electronic Journal}.

\leavevmode\vadjust pre{\hypertarget{ref-reyDilemmaNotTrilemma2015}{}}%
Rey, H. 2015. Dilemma not {Trilemma}: {The Global Financial Cycle} and
{Monetary Policy Independence}. \emph{NBER Working Papers}. No. 21162.

\leavevmode\vadjust pre{\hypertarget{ref-sala-i-martinDeterminantsLongTermGrowth2004}{}}%
Sala-I-Martin, X., Doppelhofer, G. \& Miller, R.I. 2004. Determinants of
{Long}-{Term Growth}: {A Bayesian Averaging} of {Classical Estimates}
({BACE}) {Approach}. \emph{The American Economic Review}. 94(4).

\leavevmode\vadjust pre{\hypertarget{ref-schneiderCatchingGrowthDeterminants2012}{}}%
Schneider, U. \& Wagner, M. 2012. Catching {Growth Determinants} with
the {Adaptive Lasso}: {Lassoing Growth Determinants}. \emph{German
Economic Review}. 13(1):71--85.

\leavevmode\vadjust pre{\hypertarget{ref-simonArchitectureComplexity1962}{}}%
Simon, H.A. 1962. The {Architecture} of {Complexity}. \emph{Proceedings
of the American Philosophical Society}. 106(6):467--482.

\leavevmode\vadjust pre{\hypertarget{ref-stockDynamicFactorModels2016}{}}%
Stock \& Watson. 2016a. Dynamic {Factor Models}, {Factor}-{Augmented
Vector Autoregressions}, and {Structural Vector Autoregressions} in
{Macroeconomics}. in \emph{Handbook of {Macroeconomics}} Vol. 2.
{Elsevier}. 415--525.

\leavevmode\vadjust pre{\hypertarget{ref-stockFactorModelsStructural2016}{}}%
Stock \& Watson. 2016b. Factor {Models} and {Structural Vector
Autoregressions} in {Macroeconomics}. \emph{Handbook of Macroeconomics}.
2.

\leavevmode\vadjust pre{\hypertarget{ref-tibshiraniRegressionShrinkageSelection1996}{}}%
Tibshirani, R. 1996. Regression {Shrinkage} and {Selection Via} the
{Lasso}. \emph{Journal of the Royal Statistical Society: Series B
(Methodological)}. 58(1):267--288.

\leavevmode\vadjust pre{\hypertarget{ref-tibshiraniSparsitySmoothnessFused2005}{}}%
Tibshirani, R., Saunders, M., Rosset, S., Zhu, J. \& Knight, K. 2005.
Sparsity and {Smoothness} via the {Fused Lasso}. \emph{Journal of the
Royal Statistical Society: Series B (Statistical Methodology)}.
67(1):91--108.

\leavevmode\vadjust pre{\hypertarget{ref-tolaClusterAnalysisPortfolio2008}{}}%
Tola, V., Lillo, F., Gallegati, M. \& Mantegna, R.N. 2008. Cluster
{Analysis} for {Portfolio Optimization}. \emph{Journal of Economic
Dynamics and Control}. 32(1):235--258.

\leavevmode\vadjust pre{\hypertarget{ref-tumminelloCorrelationHierarchiesNetworks2010}{}}%
Tumminello, M., Lillo, F. \& Mantegna, R.N. 2010. Correlation,
{Hierarchies}, and {Networks} in {Financial Markets}. \emph{Journal of
Economic Behavior \& Organization}. 75(1):40--58.

\leavevmode\vadjust pre{\hypertarget{ref-turlachSimultaneousVariableSelection2005}{}}%
Turlach, B.A., Venables, W.N. \& Wright, S.J. 2005. Simultaneous
{Variable Selection}. \emph{Technometrics}. 47(3):349--363.

\leavevmode\vadjust pre{\hypertarget{ref-vanwieringenLectureNotesRidge2020}{}}%
van Wieringen, W.N. 2020. \emph{Lecture {Notes} on {Ridge Regression}}.
(Paper 1509.09169). {arXiv.org}.

\leavevmode\vadjust pre{\hypertarget{ref-varianBigDataNew2014}{}}%
Varian, H.R. 2014. Big {Data}: {New Tricks} for {Econometrics}.
\emph{Journal of Economic Perspectives}. 28(2):3--28.

\leavevmode\vadjust pre{\hypertarget{ref-wardHierarchical1963}{}}%
Ward, J.H. 1963. Hierarchical grouping to optimize an objective
function. \emph{Journal of the American Statistical Association}.
58(301):236--244.

\leavevmode\vadjust pre{\hypertarget{ref-woldPersonalMemoriesEarly2001}{}}%
Wold, S. 2001. Personal {Memories} of the {Early PLS Development}.
\emph{Chemometrics and Intelligent Laboratory Systems}. 58(2):83--84.

\leavevmode\vadjust pre{\hypertarget{ref-yuanModelSelectionEstimation2006}{}}%
Yuan, M. \& Lin, Y. 2006. Model {Selection} and {Estimation} in
{Regression} with {Grouped Variables}. \emph{Journal of the Royal
Statistical Society: Series B (Statistical Methodology)}. 68(1):49--67.

\leavevmode\vadjust pre{\hypertarget{ref-zengNovelSparsityClustering2013}{}}%
Zeng, X. \& Figueiredo, M.A.T. 2013. \emph{A {Novel Sparsity} and
{Clustering Regularization}}. (Paper 1310.4945). {arXiv.org}.

\leavevmode\vadjust pre{\hypertarget{ref-zouAdaptiveLassoIts2006}{}}%
Zou, H. 2006. The {Adaptive Lasso} and {Its Oracle Properties}.
\emph{Journal of the American Statistical Association}.
101(476):1418--1429.

\leavevmode\vadjust pre{\hypertarget{ref-zouRegularizationVariableSelection2005}{}}%
Zou, H. \& Hastie, T. 2005. Regularization and {Variable Selection} via
the {Elastic Net}. \emph{Journal of the Royal Statistical Society:
Series B (Statistical Methodology)}. 67(2):301--320.

\leavevmode\vadjust pre{\hypertarget{ref-zouAdaptiveElasticNetDiverging2009}{}}%
Zou, H. \& Zhang, H.H. 2009. On the {Adaptive Elastic}-{Net} with a
{Diverging Number} of {Parameters}. \emph{The Annals of Statistics}.
37(4):1733--1751.

\end{CSLReferences}

\newpage
\appendix

\hypertarget{proof-of-proposition}{%
\section{\texorpdfstring{Proof of Proposition \ref{prop:hfr}
\label{appa}}{Proof of Proposition  }}\label{proof-of-proposition}}

Proposition \ref{prop:hfr} can be shown to hold by demonstrating the
equivalency to ordinary least squares coefficients. The proposition
defines \begin{equation}
\boldsymbol{\hat{\beta}} = \mathbf{S}^\top\mathbf{Q}_{zz}^{-1}\mathbf{Q}_{zy}.
\end{equation} Expanding the regression equation and calculating the
inverse product results in \begin{align}
\boldsymbol{\hat{\beta}} &= \begin{bmatrix}
\mathbf{S}_1 \\
\mathbf{S}_2 \\
\mathbf{S}_3
\end{bmatrix}^\top
\begin{bmatrix}
\mathbf{Q}_{11} & \boldsymbol{0} & \boldsymbol{0} \\
\mathbf{Q}_{21} & \mathbf{Q}_{22} & \boldsymbol{0} \\
\mathbf{Q}_{31} & \mathbf{Q}_{32} & \mathbf{Q}_{33}
\end{bmatrix}^{-1}
\begin{bmatrix}
\mathbf{Q}_{1y} \\
\mathbf{Q}_{2y} \\
\mathbf{Q}_{3y}
\end{bmatrix} \label{eq:appa_intro} \\
&= 
\begin{bmatrix}
\mathbf{S}_1 \\
\mathbf{S}_2 \\
\mathbf{S}_3
\end{bmatrix}^\top
\begin{bmatrix}
\mathbf{Q}_{11}^{-1} & \boldsymbol{0} & \boldsymbol{0} \\
-\mathbf{Q}_{22}^{-1}\mathbf{Q}_{21}\mathbf{Q}_{11}^{-1} & \mathbf{Q}_{22}^{-1} & \boldsymbol{0} \\
\mathbf{Q}_{33}^{-1}\mathbf{Q}_{32}\mathbf{Q}_{22}^{-1}\mathbf{Q}_{21}\mathbf{Q}_{11}^{-1}
-\mathbf{Q}_{33}^{-1}\mathbf{Q}_{31}\mathbf{Q}_{11}^{-1} &
-\mathbf{Q}_{33}^{-1}\mathbf{Q}_{32}\mathbf{Q}_{22}^{-1} &
\mathbf{Q}_{33}^{-1}
\end{bmatrix}
\begin{bmatrix}
\mathbf{Q}_{1y} \\
\mathbf{Q}_{2y} \\
\mathbf{Q}_{3y}
\end{bmatrix} \label{eq:appa_intro2}\\
&= 
\begin{bmatrix}
\mathbf{S}_1 \\
\mathbf{S}_2 \\
\mathbf{S}_3
\end{bmatrix}^\top
\begin{bmatrix}
\mathbf{Q}_{11}^{-1} \mathbf{Q}_{1y} \\
\mathbf{Q}_{22}^{-1} \mathbf{Q}_{2y} -\mathbf{Q}_{22}^{-1}\mathbf{Q}_{21}\mathbf{Q}_{11}^{-1} \mathbf{Q}_{1y} \\
\mathbf{Q}_{33}^{-1}\mathbf{Q}_{3y} - 
\mathbf{Q}_{33}^{-1}\mathbf{Q}_{32}\mathbf{Q}_{22}^{-1}\mathbf{Q}_{2y} - \mathbf{Q}_{33}^{-1}\mathbf{Q}_{31}\mathbf{Q}_{11}^{-1}\mathbf{Q}_{1y} + \mathbf{Q}_{33}^{-1}\mathbf{Q}_{32}\mathbf{Q}_{22}^{-1}\mathbf{Q}_{21}\mathbf{Q}_{11}^{-1}\mathbf{Q}_{1y}
\end{bmatrix}.
\label{eq:appa_intro3}
\end{align} Here \(\mathbf{Q}_{ij} = \mathbf{z}_i^\top\mathbf{z}_j\) and
\(\mathbf{Q}_{iy} = \mathbf{z}_i^\top y\). Using the definition of the
projection matrix
\(\mathbf{P}_i = \mathbf{z}_i\mathbf{Q}_{ii}^{-1}\mathbf{z}_i^\top\),
and substituting the definitions of \(\mathbf{Q}_{ij}\) and
\(\mathbf{Q}_{iy}\) the above can be simplified to give \begin{align}
\boldsymbol{\hat{\beta}} &= 
\begin{bmatrix}
\mathbf{S}_1 \\
\mathbf{S}_2 \\
\mathbf{S}_3
\end{bmatrix}^\top
\begin{bmatrix}
\mathbf{Q}_{11}^{-1} \mathbf{Q}_{1y} \\
\mathbf{Q}_{22}^{-1} \mathbf{z}_2[\mathbf{I} - \mathbf{P}_{1}]y \\
\mathbf{Q}_{33}^{-1}\mathbf{z}_3[\mathbf{I} - \mathbf{P}_2 - \mathbf{P}_1 + \mathbf{P}_2\mathbf{P}_1]y
\end{bmatrix} \\
&= \begin{bmatrix}
\mathbf{S}_1 \\
\mathbf{S}_2 \\
\mathbf{S}_3
\end{bmatrix}^\top
\begin{bmatrix}
\mathbf{Q}_{11}^{-1} \mathbf{Q}_{1y} \\
\mathbf{Q}_{22}^{-1} \mathbf{z}_2[\mathbf{I} - \mathbf{P}_{1}]y \\
\mathbf{Q}_{33}^{-1}\mathbf{z}_3[\mathbf{I} - \mathbf{P}_2][\mathbf{I} - \mathbf{P}_1]y
\end{bmatrix}.
\label{eq:app_stacked}
\end{align} Eq. \ref{eq:app_stacked} stacks the level-specific
estimates, where the preceding levels are partialled out of each
respective level-specific estimate. Note also that the nested nature of
the hierarchical features implies that
\([\mathbf{I} - \mathbf{P}_2][\mathbf{I} - \mathbf{P}_1]y = [\mathbf{I} - \mathbf{P}_2]y\),
making the above exactly analogous to a stacked version of Eq.
\ref{eq:level_specific_parameters}. Multiplying the matrix and using the
simple trick that
\(\mathbf{S}_i^\top(\mathbf{z}_i^\top\mathbf{z}_i)^{-1}\mathbf{z}_i^\top\mathbf{z}_j = \mathbf{S}_j^\top \; \forall \; i>j\)
allows Eq. \ref{eq:app_stacked} to be simplified further, resulting in
\begin{align}
\boldsymbol{\hat{\beta}} = \mathbf{S}_1^\top\mathbf{Q}_{11}^{-1}\mathbf{Q}_{1y} +
\mathbf{S}_2^\top\mathbf{Q}_{22}^{-1}\mathbf{Q}_{2y} - \mathbf{S}_1^\top\mathbf{Q}_{11}^{-1}\mathbf{Q}_{1y}& + 
\mathbf{S}_3^\top\mathbf{Q}_{33}^{-1}\mathbf{Q}_{3y} -  \nonumber \\
&\mathbf{S}_2^\top\mathbf{Q}_{22}^{-1}\mathbf{Q}_{2y} -
\mathbf{S}_1^\top\mathbf{Q}_{11}^{-1}\mathbf{Q}_{1y} +
\mathbf{S}_1^\top\mathbf{Q}_{11}^{-1}\mathbf{Q}_{1y}.
\end{align} Finally, recalling that \(\mathbf{S}_3 = \mathbf{I}\), this
simplifies to \begin{equation}
\boldsymbol{\hat{\beta}} = \mathbf{Q}_{33}^{-1}\mathbf{Q}_{3y}.
\end{equation} Since \(\mathbf{z}_3 = \mathbf{x}\),
\(\boldsymbol{\hat{\beta}}\) is simply the ordinary least squares
estimator: \begin{equation}
\boldsymbol{\hat{\beta}} = (\mathbf{x}^\top\mathbf{x})^{-1}\mathbf{x}^\top y = \boldsymbol{\hat{\beta}}_{\text{ols}}.
\end{equation}

\newpage

\hypertarget{proof-of-proposition-1}{%
\section{\texorpdfstring{Proof of Proposition
\ref{prop:shrinkage_vector}
\label{appb}}{Proof of Proposition  }}\label{proof-of-proposition-1}}

Let \(\nu_{\text{eff}}\) be the effective model degrees of freedom of
the HFR estimator, with \[
\nu_{\text{eff}} = \text{tr}(\mathbf{P}_{\text{hfr}})\;\;\; \text{and}\;\;\; \mathbf{P}_{\text{hfr}} = \mathbf{z}(\mathbf{z}^\top\mathbf{z}\odot\mathbf{H}\odot\mathbf{\Theta})^{-1}\mathbf{z}^\top.
\] Using Eq. \ref{eq:chain_w_shrinkage}, the HFR model fit can be
written as \begin{align}
\hat{y} &= \sum_{\ell=1}^L \theta_{\ell}\mathbf{x}\mathbf{\hat{b}}_{\ell}\\
&=\sum_{\ell=1}^L \theta_{\ell}\mathbf{xS}^\top(\mathbf{z}_{\ell}^\top\mathbf{z}_{\ell})^{-1}\mathbf{z}_{\ell}^\top\mathbf{M}_{\ell-1}y \\
&= \sum_{\ell=1}^L \theta_{\ell}\mathbf{P}_{\ell}\mathbf{M}_{\ell-1}y. \label{eq:app_yfit}
\end{align} The projection matrix of the HFR estimator can now be
rewritten as \begin{align}
\mathbf{P}_{\text{hfr}} &= \sum_{\ell=1}^L\theta_{\ell}\mathbf{P}_{\ell}\mathbf{M}_{\ell-1} \\
&= \sum_{\ell=1}^L\theta_{\ell}\mathbf{P}_{\ell}(\mathbf{I}_N - \mathbf{P}_{\ell-1}) \\
&= \sum_{\ell=1}^L\theta_{\ell}(\mathbf{P}_{\ell} -\mathbf{P}_{\ell} \mathbf{P}_{\ell-1}).
\end{align} Here \(\mathbf{M}_0 = \mathbf{I}_N\) and
\(\mathbf{P}_0 = \mathbf{0}\). Recall that for the nested case, where
each level contains strictly more information than the preceding level,
\(\mathbf{M}_{\ell-1} \equiv \prod_{i=1}^{\ell}\mathbf{M}_{\ell-i}\).
This implies that
\(\mathbf{M}_{\ell}\mathbf{M}_{\ell-1} = \mathbf{M}_{\ell}\), and by
expanding the equality,
\(\mathbf{P}_{\ell}\mathbf{P}_{\ell-1} = \mathbf{P}_{\ell-1}\).

Substituting and using the properties of the trace operator, the
effective degrees of freedom becomes \begin{equation}
\text{tr}(\mathbf{P}_{\text{hfr}}) = \sum_{\ell=1}^L\theta_{\ell}\left[\text{tr}(\mathbf{P}_{\ell}) -\text{tr}(\mathbf{P}_{\ell-1})\right].
\end{equation} With a total of \(L = K\) levels, the number of features
contained in the \(\ell\)th level --- and thus the rank of
\(\mathbf{P}_{\ell}\) --- is simply \(\ell\). The above therefore
simplifies to \begin{align}
\text{tr}(\mathbf{P}_{\text{hfr}}) &= \sum_{\ell=1}^L\theta_{\ell}\left[\ell - (\ell-1)\right] \\
&= \sum_{\ell=1}^L\theta_{\ell}\cdot 1 \\
&= \sum_{\ell=1}^L\theta_{\ell}.
\end{align}

\newpage

\hypertarget{derivation-of-path-indepdentent-hfr-estimates}{%
\section{\texorpdfstring{Derivation of path-indepdentent HFR estimates
\label{appc}}{Derivation of path-indepdentent HFR estimates }}\label{derivation-of-path-indepdentent-hfr-estimates}}

Section \ref{method:approximation} suggests that
\(\boldsymbol{\hat{\beta}}_{\text{hfr}}\) can be reformulated to remove
path-dependence from the level-specific estimates, with \begin{equation}
\boldsymbol{\hat{\beta}}_{\text{hfr}} = \boldsymbol{\hat{\mathcal{B}}}\boldsymbol{\phi}.
\end{equation} To derive this result, recall once again the case with
\(K=4\) and \(L=3\) presented in Section \ref{method:framework} and in
\ref{appa}. Here
\(\boldsymbol{\hat{\mathcal{B}}} = \begin{bmatrix} \mathbf{\hat{w}}_1 & \cdots & \mathbf{\hat{w}}_L \end{bmatrix}\),
with
\(\mathbf{\hat{w}}_{\ell} = \mathbf{S}_{\ell}^\top\mathbf{Q}_{\ell\ell}^{-1}\mathbf{Q}_{\ell y}\),
using the notation in \ref{appa}. In addition, \(\boldsymbol{\phi}\) is
the transformation of the vector of shrinkage weights described in
Section \ref{method:approximation}.

Eq. \ref{eq:appb_add_shrinkage} begins by restating Eq.
\ref{eq:appa_intro} with shrinkage weights: \begin{align}
\boldsymbol{\hat{\beta}}_{\text{hfr}} &= \begin{bmatrix}
\mathbf{S}_1 \\
\mathbf{S}_2 \\
\mathbf{S}_3
\end{bmatrix}^\top
\begin{bmatrix}
\begin{bmatrix}
\mathbf{Q}_{11} & \boldsymbol{0} & \boldsymbol{0} \\
\mathbf{Q}_{21} & \mathbf{Q}_{22} & \boldsymbol{0} \\
\mathbf{Q}_{31} & \mathbf{Q}_{32} & \mathbf{Q}_{33}
\end{bmatrix}\odot
\begin{bmatrix}
\boldsymbol{\Theta}_1 &  \boldsymbol{\Theta}_2 & \boldsymbol{\Theta}_3
\end{bmatrix}
\end{bmatrix}^{-1}
\begin{bmatrix}
\mathbf{Q}_{1y} \\
\mathbf{Q}_{2y} \\
\mathbf{Q}_{3y}
\end{bmatrix} \label{eq:appb_add_shrinkage} \\
&= \begin{bmatrix}
\mathbf{S}_1 \\
\mathbf{S}_2 \\
\mathbf{S}_3
\end{bmatrix}^\top
\begin{bmatrix}
\theta_1^{-1} \mathbf{Q}_{11} & \boldsymbol{0} & \boldsymbol{0} \\
\theta_1^{-1} \mathbf{Q}_{21} & \theta_2^{-1} \mathbf{Q}_{22} & \boldsymbol{0} \\
\theta_1^{-1} \mathbf{Q}_{31} & \theta_2^{-1} \mathbf{Q}_{32} & \theta_3^{-1} \mathbf{Q}_{33}
\end{bmatrix}^{-1}
\begin{bmatrix}
\mathbf{Q}_{1y} \\
\mathbf{Q}_{2y} \\
\mathbf{Q}_{3y}
\end{bmatrix}.
\end{align} Calculating the inverse and multiplying out in a manner
analogous to Eqs. \ref{eq:appa_intro2} \& \ref{eq:appa_intro3} in
\ref{appa} reduces the above to \begin{align}
\boldsymbol{\hat{\beta}}_{\text{hfr}} = 
\theta_1\mathbf{S}_1^\top\mathbf{Q}_{11}^{-1}\mathbf{Q}_{1y} +
\theta_2\mathbf{S}_2^\top\mathbf{Q}_{22}^{-1}\mathbf{Q}_{2y} &- \theta_2\mathbf{S}_1^\top\mathbf{Q}_{11}^{-1}\mathbf{Q}_{1y} + 
\theta_3\mathbf{S}_3^\top\mathbf{Q}_{33}^{-1}\mathbf{Q}_{3y} -  \nonumber \\
&\theta_3\mathbf{S}_2^\top\mathbf{Q}_{22}^{-1}\mathbf{Q}_{2y} -
\theta_3\mathbf{S}_1^\top\mathbf{Q}_{11}^{-1}\mathbf{Q}_{1y} +
\theta_3\mathbf{S}_1^\top\mathbf{Q}_{11}^{-1}\mathbf{Q}_{1y}.
\end{align} Using the definition of \(\mathbf{\hat{w}}_{\ell}\) yields
\begin{align}
\boldsymbol{\hat{\beta}}_{\text{hfr}} &= 
\mathbf{\hat{w}}_1 (\theta_1 - \theta_2) + \mathbf{\hat{w}}_2(\theta_2 - \theta_3) + \mathbf{\hat{w}}_3(\theta_3) \\
 &= \boldsymbol{\hat{\mathcal{B}}} \boldsymbol{\phi},
\end{align} where \begin{equation*}
\boldsymbol{\phi} = \begin{cases}
\theta_{\ell} - \theta_{\ell+1} &\text{when }\ell < 3\\
\theta_{\ell} &\text{otherwise.}
\end{cases}
\end{equation*} With the addition of an arbitrary number of levels, this
result generalizes to the definition presented in Section
\ref{method:approximation}.

\newpage

\hypertarget{description-of-growth-determinants-data-set}{%
\section{Description of growth determinants data
set}\label{description-of-growth-determinants-data-set}}

\begin{table}[ht]
\centering
\scalebox{0.7}{
\begin{tabular}{lc|lc}
  \toprule
Description & Name & Description & Name \\ 
  \midrule
Absolute Latitude & ABSLATIT & Fraction of Land Area Near Navigable Water & LT100CR \\ 
  Air Distance to Big Cities & AIRDIST & Malaria Prevalence in 1960s & MALFAL66 \\ 
  Ethnolinguistic Fractionalization & AVELF & Fraction GDP in Mining & MINING \\ 
  British Colony Dummy & BRIT & Fraction Muslim & MUSLIM00 \\ 
  Fraction Buddhist & BUDDHA & Timing of Independence & NEWSTATE \\ 
  Fraction Catholic & CATH00 & Oil Producing Country Dummy & OIL \\ 
  Civil Liberties & CIV72 & Openess measure 1965-74 & OPENDEC1 \\ 
  Colony Dummy & COLONY & Fraction Othodox & ORTH00 \\ 
  Fraction Confucian & CONFUC & Fraction Speaking Foreign Language & OTHFRAC \\ 
  Population Density 1960 & DENS60 & Primary Schooling in 1960 & P60 \\ 
  Population Density Coastal in 1960s & DENS65C & Average Inflation 1960-90 & PI6090 \\ 
  Interior Density & DENS65I & Square of Inflation 1960-90 & SQPI6090 \\ 
  Population Growth Rate 1960-90 & DPOP6090 & Political Rights & PRIGHTS \\ 
  East Asian Dummy & EAST & Fraction Population Less than 15 & POP1560 \\ 
  Capitalism & ECORG & Population in 1960 & POP60 \\ 
  English Speaking Population & ENGFRAC & Fraction Population Over 65 & POP6560 \\ 
  European Dummy & EUROPE & Primary Exports 1970 & PRIEXP70 \\ 
  Fertility in 1960s  & FERTLDC1 & Fraction Protestants & PROT00 \\ 
  Defense Spending Share & GDE1 & Real Exchange Rate Distortions & RERD \\ 
  GDP in 1960 (log) & GDPCH60L & Revolutions and Coups & REVCOUP \\ 
  Public Education Spending Share in GDP in 1960s  & GEEREC1 & African Dummy & SAFRICA \\ 
  Public Investment Share & GGCFD3 & Outward Orientation & SCOUT \\ 
  Nominal Govertnment GDP Share 1960s & GOVNOM1 & Size of Economy & SIZE60 \\ 
  Government Share of GDP in 1960s & GOVSH61 & Socialist Dummy & SOCIALIST \\ 
  Gov. Consumption Share 1960s & GVR61 & Spanish Colony & SPAIN \\ 
  Higher Education 1960 & H60 & Terms of Trade Growth in 1960s & TOT1DEC1 \\ 
  Religion Measure & HERF00 & Terms of Trade Ranking & TOTIND \\ 
  Fraction Hindus & HINDU00 & Fraction of Tropical Area & TROPICAR \\ 
  Investment Price & IPRICE1 & Fraction Population In Tropics & TROPPOP \\ 
  Latin American Dummy & LAAM & Fraction Spent in War 1960-90 & WARTIME \\ 
  Land Area & LANDAREA & War Particpation 1960-90 & WARTORN \\ 
  Landlocked Country Dummy & LANDLOCK & Years Open 1950-94 & YRSOPEN \\ 
  Hydrocarbon Deposits in 1993 & LHCPC & Tropical Climate Zone & ZTROPICS \\ 
  Life Expectancy in 1960 & LIFE060 &  &  \\ 
   \bottomrule
\end{tabular}
}
\caption{Description of growth determinants included in the dataset of Sala-I-Martin et al. (2004)} 
\label{tab:vars}
\end{table}

\bibliography{Tex/ref}

\end{document}